\documentclass[11pt]{article}

\usepackage{makecell}
\usepackage{xcolor}
\newcommand{\colordiff}[1]{\ifdim #1pt>0pt \textcolor{red}{#1}\else\ifdim #1pt<0pt \textcolor{blue}{#1}\else #1\fi\fi}

\usepackage{xcolor}
\usepackage{pdftexcmds}

\newcommand{\colordifff}[1]{%
    \ifnum\numexpr\pdfstrcmp{#1}{0}<0
        \textcolor{red}{#1}%
    \else
        \ifnum\numexpr\pdfstrcmp{#1}{0}>0
            \textcolor{blue}{#1}%
        \else
            #1%
        \fi
    \fi
}
\newcommand{\colordiffrev}[1]{\ifdim #1pt>0pt \textcolor{blue}{#1}\else\ifdim #1pt<0pt \textcolor{red}{#1}\else #1\fi\fi}

\usepackage{microtype}
\usepackage{hyperref}
\usepackage{url}
\usepackage{booktabs}
\usepackage{amsmath}
\usepackage{graphicx}
\usepackage{multirow} 
\usepackage[most]{tcolorbox}
\usepackage{amssymb}

\definecolor{darkblue}{rgb}{0, 0, 0.5}
\hypersetup{colorlinks=true, citecolor=darkblue, linkcolor=darkblue, urlcolor=darkblue}

\usepackage[final]{acl}

\usepackage{times}
\usepackage{latexsym}

\usepackage[T1]{fontenc}

\usepackage[utf8]{inputenc}

\usepackage{microtype}

\usepackage{inconsolata}

\usepackage{graphicx}

\title{Know Your Role, Take Your Turn: 

Socially Grounded Multi-Party Dialogue Agents}

\author{
Jing Jin\textsuperscript{1},
An-Zi Yen\textsuperscript{2},
Hen-Hsen Huang\textsuperscript{3},
Hsin-Hsi Chen\textsuperscript{1,4} \\
\textsuperscript{1}Department of Computer Science and Information Engineering, National Taiwan University, Taiwan \\
\textsuperscript{2}Department of Computer Science, National Yang Ming Chiao Tung University, Taiwan \\
\textsuperscript{3}Institute of Information Science, Academia Sinica, Taiwan \\
\textsuperscript{4}AI Research Center (AINTU), National Taiwan University, Taiwan \\
\texttt{cching@nlg.csie.ntu.edu.tw}, \texttt{azyen@nycu.edu.tw}, \\
\texttt{hhhuang@iis.sinica.edu.tw}, \texttt{hhchen@ntu.edu.tw}
}

\begin{document}
\maketitle
\begin{abstract}

As intelligent assistants become more deeply embedded in everyday life, future AI agents must participate naturally in dynamic multi-party conversations. Doing so requires understanding social roles, recognizing whom an utterance targets, deciding when to speak, and responding appropriately. In this work, we investigate how \textbf{social relationship information} shapes these capabilities through a \textbf{three-stage diagnostic framework} consisting of addressee recognition, turn-taking judgement, and response generation. We evaluate four LLMs on MPDD and a reconstructed English subset of Bazinga! across three levels of social information: none, positional, and relational. We additionally examine results by addressee mention type to determine when social cues provide the greatest benefit. Results show that relationship-aware prompting consistently improves addressee recognition in Chinese multi-party dialogue, with particularly clear benefits for role-based references, and can also improve turn-taking judgement and response generation. These findings suggest that socially grounded reasoning is an important component of AI agents that can better understand their roles, take appropriate turns, and participate naturally in multi-party conversation.
\footnote{Source code for reproducing all experiments is released at~\url{https://github.com/jingjin412/KYR-TYT/}.}
\end{abstract}

\section{Introduction}
As LLM-based assistants become increasingly embedded in everyday interactions, they must understand not only what is said, but also how to participate appropriately. In multi-party conversations, this requires reasoning about social roles, conversational targets, and turn-taking expectations: an agent may interrupt an exchange that does not involve it, remain silent when its participation is expected, or respond inconsistently with the relationships among participants. Such failures can disrupt interaction and violate social expectations.

This challenge is especially pronounced when participation cues are left implicit. In natural dialogue, people often rely on role expressions, family terms, hierarchical positions, and interactional expectations shaped by interpersonal relationships rather than direct naming. Interpreting such exchanges requires more than linguistic understanding: an agent must infer the social structure of the conversation to determine whom an utterance targets, who should speak next, and how to respond.

Despite its practical importance, socially grounded participation remains underexplored in the LLM era, particularly in multi-party settings where relationships shape how people address and take turns with one another. Prior work often studies individual capabilities such as addressee recognition or response generation in isolation, leaving unclear how social information supports the process of participation. For a dialogue agent, these decisions are connected: it must infer whom an utterance targets, judge whether it should speak next, and produce a response consistent with its role and relationships. Examining these capabilities together allows us to assess how social relationship information shapes multi-party interaction.

In this work, we study how social relationship information affects a three-stage dialogue-agent diagnostic framework consisting of addressee recognition, turn-taking judgement, and response generation. Since no dataset is explicitly designed for this purpose, we leverage the Multi-Party Dialogue Dataset (MPDD)~\citep{chen2020mpdd}, a Chinese corpus of TV drama scripts annotated with speakers, addressees, and interpersonal relationships, and reconstruct a smaller English corpus from the Bazinga! dataset~\citep{madotto2022bazinga} for cross-lingual evaluation. We compare conditions with no social information, coarse positional information, and fine-grained relationship information, and examine how these cues influence model behavior across datasets and model families.

\begin{figure*}[t]
    \centering
    \includegraphics[width=0.94\linewidth]{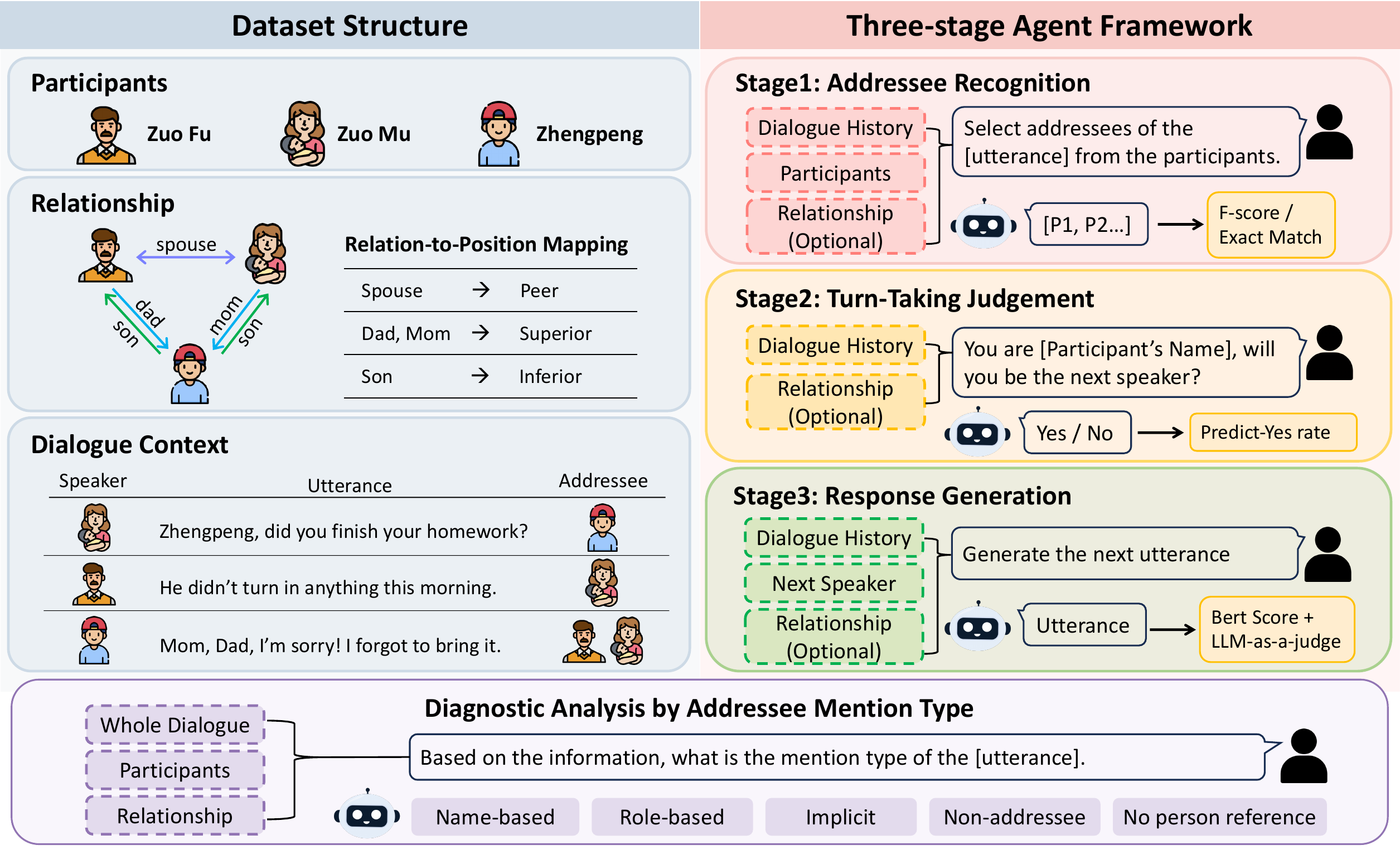}
    \caption{Overview of the proposed three-stage diagnostic framework for multi-party dialogue agents.}
    \label{fig:framework}
\end{figure*}

Our results show that relationship-aware prompting consistently improves addressee recognition on Chinese multi-party dialogue and can also benefit downstream turn-taking and response generation, although the magnitude of the effect varies by model and language. To better interpret these effects, we further analyze performance by addressee mention type, revealing when relationship information is most useful.

The contributions of this paper are summarized as follows:
\begin{itemize}
    \item We introduce a three-stage diagnostic framework for studying how social information affects multi-party dialogue understanding.
    \item We provide a controlled study of how \textbf{social information}, including coarse positional cues and fine-grained relationship cues, affects model behavior.
    \item We evaluate the framework on Chinese MPDD and reconstructed English \textit{Bazinga!}, enabling controlled cross-lingual analysis of socially aware dialogue understanding.
    \item We introduce addressee mention type annotation as a diagnostic analysis tool for characterizing when social information is most helpful.
\end{itemize}

\section{Related Work}

\subsection{Addressee Recognition}
Addressee recognition is a core challenge in multi-party dialogue because the system must infer who an utterance is directed to rather than assuming a fixed interlocutor. Early work formulated this problem jointly with response selection and showed that explicitly modeling speaker-listener interactions improves performance in multi-party settings \citep{ouchi2016addressee,zhang2018sirnn}. Later work extended addressee prediction beyond the final turn to utterance-level inference across complete conversations, highlighting the importance of modeling participant interactions throughout the dialogue \citep{le2019addressee}. Structure-aware pretraining and graph-based fine-tuning further strengthened this line of research: MPC-BERT encodes reply-to and discourse structure for multi-party understanding \citep{gu2021mpcbert}, while GIFT injects interaction graphs into transformer fine-tuning \citep{gu2023gift}. On the data side, MPDD is especially valuable because it provides speaker, addressee, emotion, and interpersonal-relation annotations in Chinese \citep{chen2020mpdd}, whereas Bazinga! offers English multi-party dialogue with speaker and addressee annotations but without an explicit relationship schema, making it useful for cross-lingual comparison but less directly suitable for socially grounded analysis \citep{madotto2022bazinga}.

\subsection{Social and Structural Cues for Multi-Party Reasoning}
Beyond addressee recognition alone, prior work has emphasized that multi-party dialogue understanding depends on social and structural cues such as reply-to relations, participant roles, and interpersonal dependencies. For example, reply-to structure has been shown to improve dialogue discourse parsing and to provide useful supervision for identifying conversational targets \citep{wang2024replytostructure}. Related work on persona- and relation-aware modeling likewise suggests that social information can improve contextual appropriateness in multi-party dialogue \citep{mahajan2024persona,lee2025mirror}. These findings collectively suggest that multi-party systems benefit from going beyond local lexical matching and instead reasoning over socially grounded interaction structure. Our work follows this direction, but differs in explicitly comparing coarse status cues and fine-grained relationship cues within a unified three-stage diagnostic framework for multi-party dialogue agents.

\subsection{LLMs for Multi-Party Dialogue}
Recent studies show that large language models remain far from solved on multi-party dialogue. Zero-shot evaluations found that ChatGPT and GPT-4 can handle some multi-party tasks, but performance is still sensitive to how interaction structure is presented in the prompt \citep{tan2023evaluations}. Diagnostic analyses further show that LLMs are more brittle on addressee recognition than on response selection, especially when addressing is implicit or when conversational structure is not made explicit \citep{penzo2024hangover}. This limitation has also motivated multimodal benchmarks, which demonstrate that addressee reasoning may depend on visual and auditory cues in addition to text \citep{inoue2025benchmark}. In parallel, recent work has begun to connect addressee reasoning to downstream generation, either by incorporating latent addressee information into response generation \citep{li2023empretrain}, optimizing topic and rhetorical coherence in multi-party responses \citep{fan2024rltrc}, or fine-tuning language models directly for multi-party generation and next-speaker prediction \citep{wang2024mupas}. Compared with these studies, our work focuses on how explicit social relationship information affects not only addressee recognition, but also turn-taking judgement and response generation in a single evaluation framework.

\section{Methodology}
\label{headings}

Figure~\ref{fig:framework} illustrates our overall framework.  
We study how \textbf{social information} affects a \textbf{three-stage diagnostic framework} in multi-party dialogue. 
The framework consists of: 
(1) \textbf{Addressee Recognition}, where the model identifies the intended addressee(s) of the current utterance; 
(2) \textbf{Turn-taking Judgement}, where the model, assigned the identity of each participant, judges whether it could naturally be the next speaker; and 
(3) \textbf{Response Generation}, where the model generates the next utterance conditioned on the gold next-speaker identity. 

Across all stages, the only controlled factor is the inclusion of a \textbf{social-information block}, with three settings:
$\mathcal{S}\in\{\textsc{none}, \textsc{position}, \textsc{relation}\}$.
\textsc{Position} provides coarse hierarchical categories such as \textit{superior}, \textit{peer}, and \textit{inferior}, while \textsc{Relation} provides fine-grained interpersonal relations (e.g., \textit{parent--child}, \textit{spouse}), which can be mapped to the corresponding position categories. (Appendix \ref{sec:mapping})

\subsection{Task Setup}
We model a dialogue as a sequence of utterances
$D=\{u_1,\dots,u_m\}$,
where each utterance $u_i=(s_i,c_i)$ consists of a speaker $s_i$ and content $c_i$.
Let $\mathcal{C}$ denote the participant set of the dialogue.
For each target turn $u_i$, the model is given only the \textbf{preceding dialogue context} $D_{\le i}$ to simulate real-time conversational assistance.

Prompts may include the participant list and optional social information in addition to the current utterance.
The framework evaluates whether structured social cues improve not only addressee recognition but also downstream turn-taking and response generation.

Unless otherwise specified, all main experiments use a fixed prompt template for each stage, so that the only controlled variable is the level of social information (\textsc{none}, \textsc{position}, or \textsc{relation}).
To examine prompt robustness, we additionally evaluate multiple prompt rewrites generated by LLMs; these results are reported in the Appendix \ref{app: prompt sensitivity}.

\subsection{Stage 1: Addressee Recognition}
In the first stage, the model predicts the set of intended addressees for the target utterance $u_i$.
This is formulated as a multi-label prediction problem:
\[
\hat{\mathcal{L}}_i = f^{(1)}_\theta(D_{\le i},\, \mathcal{C},\, \mathcal{S}),
\]
where $\hat{\mathcal{L}}_i \subseteq \mathcal{C}$ denotes the predicted addressee set.

The prompt includes:
(i) the participant list,
(ii) the target utterance and its speaker,
(iii) the preceding dialogue history,
and optionally
(iv) the social-information block $\mathcal{S}$.

We evaluate this stage using \textbf{F-score} and \textbf{Exact Match accuracy}, following the standard multi-label addressee recognition setting.

\subsection{Stage 2: Turn-taking Judgement}

The second stage evaluates whether the model considers a participant plausible as the next speaker. For each target turn, we assign the model each participant role other than the current speaker. Given a candidate participant $p$, the model judges whether it could plausibly be the next person to speak:

\[
\hat{y}_{i,p} = f^{(2)}_\theta(D_{\le i}, p, \mathcal{S}) \in \{\texttt{Yes}, \texttt{No}\}.
\]

Each candidate is queried independently using only the preceding context $D_{\le i}$; the gold next utterance is never shown. We report the \textbf{gold-speaker Yes rate}, measuring whether the actual next speaker is recognized as plausible, and the \textbf{non-gold-speaker Yes rate} across all other candidate roles, reflecting the model's general tendency 
to permit participants to take the next turn.

\subsection{Stage 3: Response Generation}
In the third stage, the model is assigned the identity of the gold next speaker and asked to generate the next utterance.
Formally, conditioned on the dialogue history, next-speaker identity, and optional social information, the model generates
\[
\hat{c}_{i+1} = f^{(3)}_\theta(D_{\le i},\, s_{i+1},\, \mathcal{S}).
\]

This stage evaluates whether social information helps the model produce a response that is more compatible with the gold continuation.
We first report \textbf{BERTScore} against the reference next utterance as a semantic similarity metric.
Because lexical similarity alone cannot fully capture pragmatic appropriateness or discourse coherence, we also adopt a \textbf{two-stage LLM-as-a-judge} protocol.
The first stage evaluates the response using the preceding dialogue context, while the second additionally considers the future gold dialogue.
The evaluation covers six dimensions: \textit{Context Relevance}, \textit{Role Consistency}, \textit{Naturalness}, \textit{Appropriateness}, \textit{Future Coherence}, and \textit{Non-Contradiction}.
We use Qwen3-32B, GPT-OSS-120B, and Llama-3.3-70B-Instruct~\citep{meta2024llama33} as judges, and report the average score across them.

\subsection{Addressee Mention Type Annotation}
We additionally annotate each utterance with an \textbf{addressee mention type} for stratified analysis.
Specifically, we consider five categories:
\textit{Name-based}, where the addressee is explicitly referred to by name or nickname;
\textit{Role-based}, where the addressee is referred to by a social role or title;
\textit{Implicit}, where the utterance is directed to an addressee without an explicit referring expression;
\textit{Non-addressee reference}, where a person is mentioned but is not the intended addressee;
and \textit{No-reference}, where no participant is explicitly or implicitly referred to.
For each instance, Qwen3-32B, GPT-OSS-120B, and Meta-Llama-3.1-8B independently assign labels, and the final label is determined by majority vote, with manual adjudication applied to disputed cases.
Examples and annotation details are provided in the Appendix \ref{app: mention type}.

\section{Experiments}

\subsection{Datasets}

We conduct experiments on two corpora: the \textbf{Multi-Party Dialogue Dataset (MPDD)}~\citep{chen2020mpdd} and a reconstructed English subset of \textbf{Bazinga!}~\citep{madotto2022bazinga}.

\textbf{MPDD} is a Chinese multi-party dialogue corpus derived from TV drama scripts, where each utterance is annotated with its speaker, addressee(s), social relation, and emotion.
The social relations (e.g., \textit{parent}, \textit{spouse}, \textit{colleague}) are further grouped into three coarse hierarchical categories, namely \textit{superior}, \textit{peer}, and \textit{inferior}, to provide structured social cues without relying solely on fine-grained relation labels (Table~\ref{tab:relation-map}).
Since each utterance may target multiple addressees, the task is formulated as a multi-label prediction problem.

As shown in Table~\ref{tab:dataset-comparison}, MPDD is the only dataset in our study with explicit speaker--addressee--relation annotations.
To examine cross-lingual generalization, we additionally reconstruct a smaller English corpus from \textbf{Bazinga!}, whose scripts are drawn from well-known TV series.
For this subset, we infer character relations from external resources, apply the same relation-to-position mapping used in MPDD, and remove low-quality turns.
The final Bazinga! subset contains \textbf{246 dialogues} and \textbf{2,587 utterances}.
Although smaller than MPDD, which contains \textbf{1,033 dialogues} and \textbf{8,978 utterances}, it remains useful to validate whether overall trends generalize across languages.
Details of the reconstruction process are provided in the Appendix~\ref{tab:bazinga_construct}.

\begin{table}[t]
\centering
\small
\newcommand{\ds}[2]{#1~{\scriptsize\citep{#2}}}
\setlength{\tabcolsep}{3.8pt}
\begin{tabular}{@{}lcccc@{}}
\toprule
\textbf{Dataset} & \textbf{Lang.} & \textbf{Spk.} & \textbf{Addr.} & \textbf{Rel.} \\
\midrule
\ds{MPDD}{chen2020mpdd}                & Zh    & \checkmark & \checkmark & \checkmark \\
\ds{CRECIL}{jiang-etal-2022-crecil}          & Zh    & \checkmark &            & \checkmark \\
\ds{JY-QuotePlus}{yan-etal-2024-identifying} & Zh    & \checkmark & \checkmark &            \\
\ds{DialogRE}{yu-etal-2020-dialogue}         & En/Zh & \checkmark &            & \checkmark \\
\ds{Bazinga!}{madotto2022bazinga}      & En    & \checkmark & \checkmark &            \\
\ds{RiQuA}{papay-pado-2020-riqua}            & En    & \checkmark & \checkmark &            \\
\bottomrule
\end{tabular}
\caption{Comparison of dialogue datasets with speaker, addressee, and relation annotations.}
\label{tab:dataset-comparison}
\end{table}

\subsection{Models}

We evaluate four LLMs in the dialogue-agent framework: \textbf{GPT-5 mini} \citep{openai2025gpt5mini}, \textbf{Qwen3-32B} \citep{yang2025qwen3}, \textbf{LLaMA-3.1-8B} \citep{dubey2024llama3}, and \textbf{gpt-oss-120b} \citep{openai2025gptoss120b}. 
The model set is chosen to cover proprietary and open-weight systems, as well as variation in model scale and reasoning style, allowing us to test whether the effect of social relationship information is robust across heterogeneous LLMs. 
Among them, \textbf{GPT-5 mini} serves as a proprietary model \citep{openai2025gpt5mini}, \textbf{Qwen3-32B} is included for its strong multilingual capabilities and suitability for the Chinese MPDD setting \citep{yang2025qwen3}, \textbf{LLaMA-3.1-8B} serves as a widely used smaller open-weight baseline \citep{dubey2024llama3}, and \textbf{gpt-oss-120b} is included as a reasoning-oriented open-weight model \citep{openai2025gptoss120b}.

\subsection{Evaluation Metrics}

For \textbf{Stage 1 (Addressee Recognition)}, we report \textbf{F1 score} and \textbf{Exact Match Accuracy}.
F1 captures partial overlap between predicted and gold addressee sets, whereas Exact Match Accuracy counts a prediction as correct only when the predicted set exactly matches the gold set.
We treat Exact Match Accuracy as the primary metric, since missing or over-predicting even one addressee may disrupt downstream interaction.
For \textbf{Stage 2 (Turn-taking Judgement)}, we report the \textbf{gold-speaker Yes rate} and the \textbf{non-gold-speaker Yes rate}, computed over the gold next-speaker roles and all queried participant roles.
For \textbf{Stage 3 (Response Generation)}, we first report \textbf{BERTScore} against the reference next utterance.
We further evaluate generated responses using a two-stage \textbf{LLM-as-a-judge} protocol, where the first stage assesses quality based on the preceding dialogue context and the second stage evaluates compatibility with the future gold dialogue.
We report the average judge score across multiple LLM judges.

\subsection{Addressee Recognition Task}

\begin{table*}[h!]
    \centering
    \setlength{\tabcolsep}{2.7pt} 
    \small
    \begin{tabular}{clcccccccccc}
    \toprule
    \multirow{3}{*}{\textbf{Dataset}} & \multirow{3}{*}{\textbf{Model}} & \multicolumn{4}{c}{\textbf{Exact Match Accuracy}} & & \multicolumn{4}{c}{\textbf{F-score}} \\
    \cmidrule(lr){3-6} \cmidrule(lr){8-11} &
    & \textbf{w/o} & \multicolumn{2}{c}{\textbf{w/ Info}} & \textbf{Diff} & & \textbf{w/o} & \multicolumn{2}{c}{\textbf{w/ Info}} & \textbf{Diff} \\
    \cmidrule(lr){4-5} \cmidrule(lr){9-10} &
    & \textbf{Info} & \textbf{Pos.} & \textbf{Rel.} & \textbf{(w/o - w/)} & & \textbf{Info} & \textbf{Pos.} & \textbf{Rel.} & \textbf{(w/o - w/)} \\
    \midrule
     \multirow{4}{*}{\textbf{MPDD}} & 
    GPT        & 0.7319 & 0.7927 & 0.7949 & \colordiff{-0.0608}$^\dag$/\colordiff{-0.0630}$^\dag$ 
    && 0.8407 & 0.9006 & 0.9032 & \colordiff{-0.0600}$^\dag$/\colordiff{-0.0625}$^\dag$ \\ &
    Qwen       & 0.7311 & 0.7393 & 0.7386 & \colordiff{-0.0081}$^\dag$/\colordiff{-0.0075}$^\dag$ 
    && 0.8793 & 0.8892 & 0.8862 & \colordiff{-0.0100}$^\dag$/\colordiff{-0.0070}$^\dag$ \\ &
    LLaMA      & 0.6142 & 0.6286 & 0.6407 & \colordiff{-0.0145}$^\dag$/\colordiff{-0.0265}$^\dag$ 
    && 0.7269 & 0.7481 & 0.7576 & \colordiff{-0.0212}$^\dag$/\colordiff{-0.0308}$^\dag$ \\ &
    GPT-OSS        & 0.7565 & 0.7762 & 0.7807 & \colordiff{-0.0197}$^\dag$/\colordiff{-0.0242}$^\dag$ 
    && 0.8640 & 0.8779 & 0.8846 & \colordiff{-0.0138}$^\dag$/\colordiff{-0.0206}$^\dag$ \\
    \midrule
    \multirow{4}{*}{\textbf{Bazinga!}} &
    GPT        & 0.7600 & 0.7638 & 0.7642 & \colordiff{-0.0039}/\colordiff{-0.0043} 
    && 0.8562 & 0.8583 & 0.8577 & \colordiff{-0.0021}/\colordiff{-0.0016} \\ &
    Qwen       & 0.6278 & 0.6691 & 0.6637 & \colordiff{-0.0414}$^\dag$/\colordiff{-0.0359}$^\dag$ 
    && 0.8160 & 0.8426 & 0.8363 & \colordiff{-0.0267}$^\dag$/\colordiff{-0.0204}$^\dag$ \\ &
    LLaMA      & 0.6424 & 0.7256 & 0.7236 & \colordiff{-0.0831}$^\dag$/\colordiff{-0.0812}$^\dag$ 
    && 0.7545 & 0.8128 & 0.8177 & \colordiff{-0.0584}$^\dag$/\colordiff{-0.0632}$^\dag$ \\ &
    GPT-OSS        & 0.7445 & 0.7812 & 0.7750 & \colordiff{-0.0367}$^\dag$/\colordiff{-0.0305}$^\dag$ 
    && 0.8356 & 0.8624 & 0.8557 & \colordiff{-0.0269}$^\dag$/\colordiff{-0.0201}$^\dag$ \\
    \bottomrule
    \end{tabular}
    \caption{Comparison of addressee prediction performance with and without participant-level social information.
    Diff = (w/o - w/) Participant Information (Position / Relation). Blue marks an improvement from social information and red a decrease. Significance levels: $^\dag p < 0.05$.
    }
    \label{tab:social-info-impact}
\end{table*}

Table \ref{tab:social-info-impact} shows the results of addressee recognition task. Across both datasets, adding participant-level social information generally improves addressee recognition, although the magnitude of the gain varies across models and datasets.

On \textbf{MPDD}, all four LLMs benefit from social information under both \textsc{Position} and \textsc{Relation} settings, and most improvements are statistically significant.
The largest gains are observed for \textbf{GPT}, whose Exact Match Accuracy increases from 0.7319 to 0.7927/0.7949 and whose F-score rises from 0.8407 to 0.9006/0.9032.
\textbf{LLaMA} and \textbf{GPT-OSS} also show clear improvements, while \textbf{Qwen} exhibits smaller but still consistent gains.
In most cases, \textsc{Relation} performs slightly better than \textsc{Position}, suggesting that finer-grained interpersonal information provides stronger cues for disambiguating addressees.

On \textbf{Bazinga!}, the effect of social information is more model-dependent.
\textbf{LLaMA}, \textbf{GPT-OSS}, and \textbf{Qwen} all show substantial improvements over the \textsc{none} setting, with especially large gains for LLaMA.
For example, LLaMA improves from 0.6424 to 0.7256/0.7236 in Exact Match Accuracy and from 0.7545 to 0.8128/0.8177 in F-score.
\textbf{GPT} shows marginal changes, and these differences aren't statistically significant.
Compared with MPDD, the benefits on Bazinga! are less uniform, although social information remains helpful for most models.

Overall, these results indicate that participant-level social information is a robust signal for addressee recognition, especially on MPDD, where the gains are consistent across all evaluated LLMs.

\paragraph{Fine-tuned LLMs.}
To test whether the benefit of social information persists beyond prompting,
we fine-tune LLaMA-3.1-8B and Qwen3-32B with (Q)LoRA under 5-fold
dialogue-level cross-validation (Appendix~\ref{app:finetune}).
Fine-tuning raises Exact Match in every setting
(e.g., 0.6142$\rightarrow$0.8089 for LLaMA on MPDD without social information),
yet social information still yields significant gains on top of it:
POSITION and RELATION improve Exact Match by 5.9--9.1 points across both
datasets and models ($p<0.05$).
On MPDD, these gains are larger than in the zero-shot setting, suggesting that models can learn to exploit
social information when trained on the task, rather than merely retrieving it
from pretrained priors.
Under fine-tuning, POSITION and RELATION perform comparably.

\subsection{Turn-Taking Judgement}

\begin{table*}[h!]
    \centering
    \setlength{\tabcolsep}{2.7pt}
    \small
    \begin{tabular}{clcccccccccc}
    \toprule
    \multirow{3}{*}{\textbf{Dataset}}
    & \multirow{3}{*}{\textbf{Model}}
    & \multicolumn{4}{c}{\textbf{Gold-speaker Yes Rate ($\uparrow$)}}
    & & \multicolumn{4}{c}{\textbf{Non-gold-speaker Yes Rate ($\downarrow$)}} \\
    \cmidrule(lr){3-6}
    \cmidrule(lr){8-11}
    & & \textbf{w/o} & \multicolumn{2}{c}{\textbf{w/ Info}} & \textbf{Diff}
    & & \textbf{w/o} & \multicolumn{2}{c}{\textbf{w/ Info}} & \textbf{Diff} \\
    \cmidrule(lr){4-5}
    \cmidrule(lr){9-10}
    & & \textbf{Info} & \textbf{Pos.} & \textbf{Rel.} & \textbf{(w/o - w/)}
    & & \textbf{Info} & \textbf{Pos.} & \textbf{Rel.} & \textbf{(w/o - w/)} \\
    \midrule
    \multirow{4}{*}{\textbf{MPDD}}
    & GPT     & 0.8063 & 0.8400 & 0.8463 & \colordiff{-0.0337}$^\dag$ / \colordiff{-0.0400}$^\dag$
              && 0.7036 & 0.6547 & 0.6644 & \colordiffrev{0.0489}$^\dag$ / \colordiffrev{0.0392}$^\dag$ \\
    & Qwen    & 0.6816 & 0.7104 & 0.7632 & \colordiff{-0.0288}$^\dag$ / \colordiff{-0.0816}$^\dag$
              && 0.5577 & 0.5757 & 0.6289 & \colordiffrev{-0.0180}$^\dag$ / \colordiffrev{-0.0712}$^\dag$ \\
    & LLaMA   & 0.4477 & 0.2641 & 0.3507 & \colordiff{0.1836}$^\dag$ / \colordiff{0.0970}$^\dag$
              && 0.4082 & 0.2233 & 0.3114 & \colordiffrev{0.1849}$^\dag$ / \colordiffrev{0.0968}$^\dag$ \\
    & GPT-OSS & 0.8855 & 0.9171 & 0.9087 & \colordiff{-0.0316}$^\dag$ / \colordiff{-0.0232}$^\dag$
              && 0.7918 & 0.8296 & 0.8085 & \colordiffrev{-0.0378}$^\dag$ / \colordiffrev{-0.0167}$^\dag$ \\
    \midrule
    \multirow{4}{*}{\textbf{Bazinga!}}
    & GPT     & 0.9543 & 0.9509 & 0.9603 & \colordiff{0.0034} / \colordiff{-0.0060}
              && 0.7873 & 0.7417 & 0.7433 & \colordiffrev{0.0456}$^\dag$ / \colordiffrev{0.0440}$^\dag$ \\
    & Qwen    & 0.6595 & 0.6416 & 0.6096 & \colordiff{0.0179} / \colordiff{0.0499}$^\dag$
              && 0.4845 & 0.4342 & 0.4124 & \colordiffrev{0.0503}$^\dag$ / \colordiffrev{0.0721}$^\dag$ \\
    & LLaMA   & 0.2648 & 0.2935 & 0.2802 & \colordiff{-0.0287}$^\dag$ / \colordiff{-0.0154}
              && 0.2446 & 0.1857 & 0.2075 & \colordiffrev{0.0589}$^\dag$ / \colordiffrev{0.0371}$^\dag$ \\
    & GPT-OSS & 0.9974 & 0.9966 & 0.9974 & \colordiff{0.0008} / \colordiff{0.0000}
              && 0.9952 & 0.9918 & 0.9908 & \colordiffrev{0.0034} / \colordiffrev{0.0044} \\
    \bottomrule
    \end{tabular}
    \caption{Gold-speaker (higher is better) and non-gold-speaker (lower is better) Yes rates for turn-taking judgement. Diff and significance notation follow Table~\ref{tab:social-info-impact}. Blue marks an improvement from social information (a higher gold-speaker rate or a lower non-gold-speaker rate), and red marks the opposite.}
    \label{tab:social-info-impact-turn-taking}
\end{table*}

Table~\ref{tab:social-info-impact-turn-taking} reports the results of
\textbf{Turn-Taking Judgement}. Unlike next-speaker prediction,
this task doesn't require the model to select a single participant.
The model is assigned each candidate role other than the current
speaker and judges whether that participant could plausibly speak next.
The gold-speaker Yes rate measures whether the actual next speaker is
recognized as a plausible continuer, the non-gold-speaker Yes rate
reflects the model's tendency to accept alternative participants.
Social information may improve turn-taking judgement by increasing
acceptance of the gold speaker, decreasing acceptance of non-gold speakers,
or both.

On \textbf{MPDD}, \textbf{GPT} shows the clearest improvement in
turn-taking discrimination: social information increases its gold-speaker
Yes rate while reducing acceptance of non-gold speakers. \textbf{Qwen}
also gains substantially on the gold speaker, especially under
\textsc{Relation}, but its non-gold Yes rate rises as well, suggesting a
broader tendency to accept candidates. \textbf{GPT-OSS} exhibits a similar but weaker pattern.
For \textbf{LLaMA}, both rates decrease by comparable amounts, indicating
more conservative judgements rather than better discrimination.

On \textbf{Bazinga!}, social information primarily helps models reject
non-gold speakers. \textbf{GPT} maintains a near-ceiling gold-speaker rate,
whereas \textbf{Qwen}'s improved rejection of non-gold speakers comes with
lower acceptance of the true next speaker. \textbf{LLaMA} shows the
clearest overall improvement, particularly under \textsc{Position}, with
a higher gold-speaker rate and a substantially lower non-gold rate.
In contrast, \textbf{GPT-OSS} assigns Yes to nearly all candidates,
revealing limited discrimination despite its near-ceiling gold-speaker
performance.

Overall, social information can improve turn-taking judgement by either
increasing acceptance of the gold speaker or reducing acceptance of
non-gold candidates.

\subsection{Response Generation}

\begin{table*}[h!]
    \centering
    \setlength{\tabcolsep}{2.7pt}
    \small
    \begin{tabular}{clccccccccc}
    \toprule
    \multirow{3}{*}{\textbf{Dataset}} &
    \multirow{3}{*}{\textbf{Model}} & \multicolumn{4}{c}{\textbf{BERTScore}} & & \multicolumn{4}{c}{\textbf{Final Weighted Score}} \\
    \cmidrule(lr){3-6} \cmidrule(lr){8-11} &
    & \textbf{w/o} & \multicolumn{2}{c}{\textbf{w/ Info}} & \textbf{Diff} & & \textbf{w/o} & \multicolumn{2}{c}{\textbf{w/ Info}} & \textbf{Diff} \\
    \cmidrule(lr){4-5} \cmidrule(lr){9-10} &
    & \textbf{Info} & \textbf{Pos.} & \textbf{Rel.} & \textbf{(w/o - w/)} & & \textbf{Info} & \textbf{Pos.} & \textbf{Rel.} & \textbf{(w/o - w/)} \\
    \midrule
    \multirow{4}{*}{\textbf{MPDD}} &
    GPT        & 0.5916 & 0.5916 & 0.5943 & \colordiff{0.0000}/\colordiff{-0.0026}$^\dag$
    && 4.0034 & 3.9144 & 4.1233 & \colordiff{0.0889}$^\dag$/\colordiff{-0.1199}$^\dag$ \\ &
    Qwen       & 0.6100 & 0.6085 & 0.6099 & \colordiff{0.0014}$^\dag$/0.0000
    && 4.1420 & 4.1257 & 4.2628 & \colordiff{0.0163}$^\dag$/\colordiff{-0.1208}$^\dag$ \\ &
    LLaMA      & 0.6093 & 0.6098 & 0.6109 & \colordiff{-0.0005}/\colordiff{-0.0016}$^\dag$
    && 3.3919 & 3.4568 & 3.4690 & \colordiff{-0.0648}$^\dag$/\colordiff{-0.0770}$^\dag$ \\ &
    GPT-OSS    & 0.5973 & 0.5952 & 0.5979 & \colordiff{0.0021}$^\dag$/\colordiff{-0.0006}
    && 4.0463 & 3.9519 & 4.1058 & \colordiff{0.0944}$^\dag$/\colordiff{-0.0595}$^\dag$ \\
    \midrule
    \multirow{4}{*}{\textbf{Bazinga!}} &
    GPT        & 0.6549 & 0.6556 & 0.6536 & \colordiff{-0.0006}/\colordiff{0.0013}$^\dag$
    && 4.1209 & 4.0914 & 4.1084 & \colordiff{0.0294}$^\dag$/\colordiff{0.0125} \\ &
    Qwen       & 0.6622 & 0.6655 & 0.6648 & \colordiff{-0.0033}$^\dag$/\colordiff{-0.0026}$^\dag$
    && 4.1400 & 4.1392 & 4.1715 & \colordiff{0.0008}/\colordiff{-0.0315}$^\dag$ \\ &
    LLaMA      & 0.6525 & 0.6563 & 0.6560 & \colordiff{-0.0038}$^\dag$/\colordiff{-0.0035}$^\dag$
    && 3.7572 & 3.7993 & 3.8285 & \colordiff{-0.0421}$^\dag$/\colordiff{-0.0712}$^\dag$ \\ &
    GPT-OSS    & 0.6464 & 0.6506 & 0.6494 & \colordiff{-0.0042}$^\dag$/\colordiff{-0.0030}$^\dag$
    && 4.1284 & 4.1553 & 4.1865 & \colordiff{-0.0269}$^\dag$/\colordiff{-0.0580}$^\dag$ \\
    \bottomrule
    \end{tabular}
    \caption{Comparison of response generation performance with and without social information.
    We report BERTScore and the judge-based Final Weighted Score (FWS), averaged over three LLM judges
    (Qwen3-32B, GPT-OSS-120B, and Llama-3.3-70B-Instruct).
    Definitions of Diff, color coding, and significance notation follow Table~\ref{tab:social-info-impact}.}
    \label{tab:response-generation}
\end{table*}

Table~\ref{tab:response-generation} presents the results of \textbf{Response Generation}.
We first evaluate each generated reply against the gold next
utterance using \textbf{BERTScore}.
However, similarity to a single reference does not necessarily
reflect whether a reply is socially appropriate or coherent
with the broader conversation.
We therefore adopt a \textbf{two-stage LLM-as-a-judge} evaluation.
The first stage assesses context relevance, role consistency,
naturalness, and appropriateness based on the dialogue history
up to the target turn.
The second stage additionally considers the subsequent gold
utterances to assess whether the generated reply remains
coherent with the unfolding conversation without introducing
contradictions.
To ensure a fair comparison across the three generation
settings, judges receive the same participant relationship
information regardless of the information available to the
generation model.

Each criterion is rated on a 1--5 scale, and the two stages
are combined into a \textbf{Final Weighted Score (FWS)}, with
weights of 60\% and 40\%, respectively.
To mitigate individual judge bias, we average the scores
across three judges: Qwen3-32B, GPT-OSS-120B, and
Llama-3.3-70B-Instruct \citep{meta2024llama33}.
The aggregated scores exhibit good inter-judge reliability
(ICC(3,$k$) = 0.82 on MPDD and 0.78 on Bazinga!;
Appendix~\ref{app:judge-agreement}) and correlate positively
with human ratings (Spearman's $\rho$ = 0.56 across 2,400
ratings collected from three annotators;
Appendix~\ref{app:human-eval}).

\textbf{BERTScore changes are small.}
Social information changes BERTScore by at most 0.0042 (under 1\%). It significantly improves BERTScore for Qwen, LLaMA, and GPT-OSS on Bazinga!, while on MPDD \textsc{Position} lowers it for Qwen and GPT-OSS and \textsc{Relation} yields only small significant gains for GPT and LLaMA. Since an appropriate reply can differ lexically from the single gold reply, we focus on the judge-based scores.

\textbf{Relation improves judged quality.}
\textsc{Relation} significantly raises FWS in seven of eight model--dataset combinations, with larger gains on MPDD (0.06--0.12 points) than on Bazinga! (0.03--0.07). For example, GPT improves from 4.00 to 4.12 and Qwen from 4.14 to 4.26 on MPDD. The only exception, GPT on Bazinga!, shows a small non-significant drop from an already high baseline. Gains in Role Consistency are even larger (0.14--0.23 points on MPDD; Appendix~\ref{app:judge-dimensions}), suggesting that relation information mainly helps models address their specific interlocutor appropriately.

\textbf{Position has mixed effects.}
\textsc{Position} significantly improves FWS in three combinations but lowers it in four; only LLaMA, the weakest generator, benefits on both datasets. Coarse status labels may lead to a register that fits the status but not the relation, such as addressing one's mother by name (Appendix~\ref{app:qualitative_cases}). Accordingly, \textsc{Relation} beats \textsc{Position} in all eight combinations.

\textbf{Human ratings support the trend on MPDD.}
Human scores correlate moderately with the judges ($\rho = 0.56$). On MPDD, \textsc{Relation} raises human scores for all four models (significant for GPT), and annotators, unlike the judges, also rate \textsc{Position} replies higher for GPT and Qwen; no difference is significant on Bazinga! (Appendix~\ref{app:human-eval}).

Overall, relationship information improves judged response quality while leaving reference similarity almost unchanged, suggesting that reference-based metrics may underestimate its effect.

\begin{figure*}
    \centering
    \includegraphics[width=\linewidth]{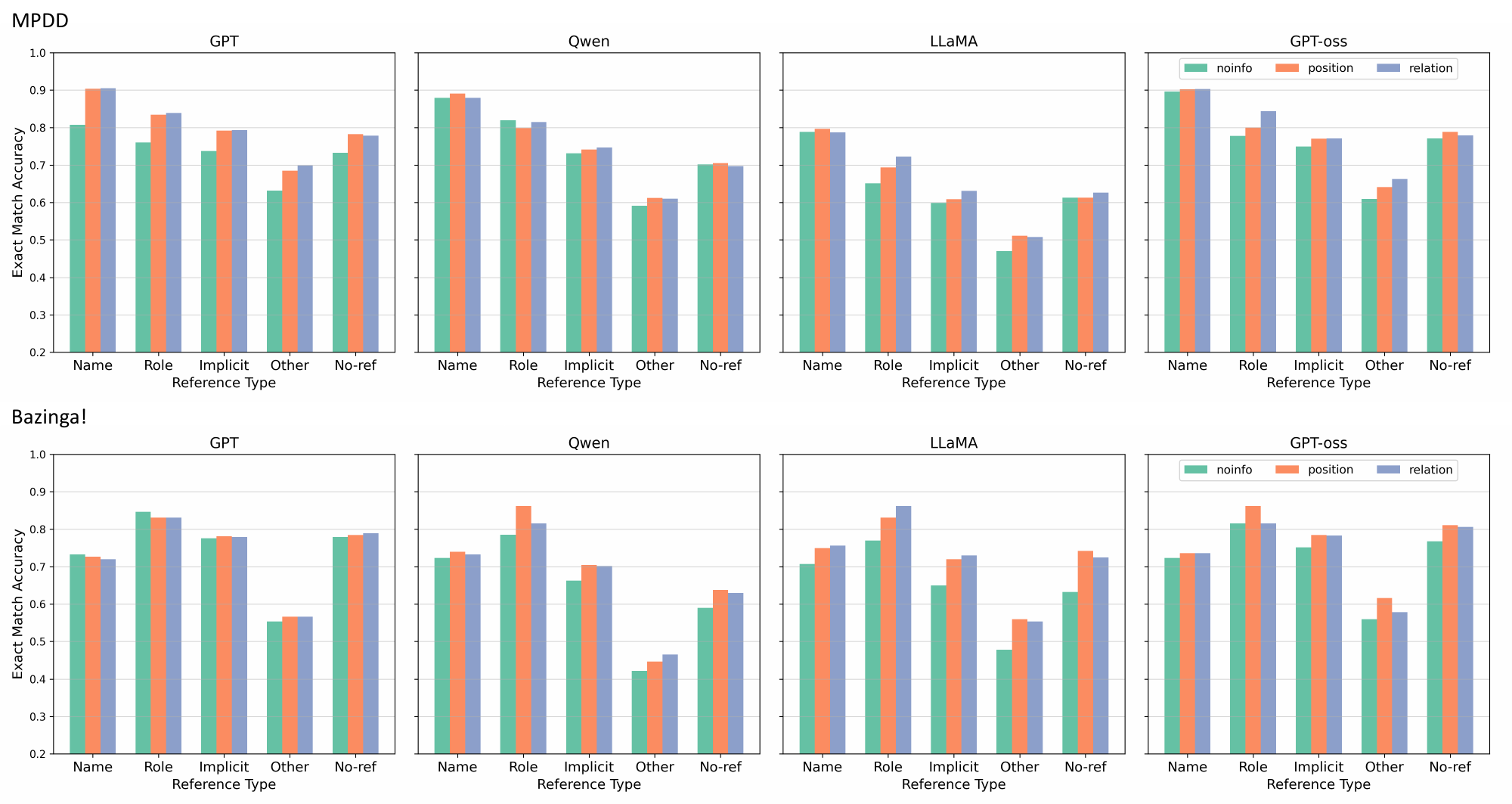}
    \caption{Exact match accuracy of addressee recognition across different reference styles and levels of participant information. All models are evaluated under the same prompt setting: using previous dialogue context and referring to participants by their real names.}
    \label{fig:ABC}
\end{figure*}

\subsection{Addressee Mention Type Analysis}

Because mention type is defined by how the current utterance refers to its intended addressee, we analyze it specifically within the \textbf{addressee recognition} task.
We consider five mention types: \textbf{name-based}, \textbf{role-based}, \textbf{implicit}, \textbf{non-addressee reference}, and \textbf{no-reference}.
To obtain these labels, we use three LLM annotators and take the majority-vote result as the final label; cases with disagreement are manually reviewed.
We adopt this LLM-centered annotation procedure because the goal of this analysis is to explain model behavior in LLM-based addressee recognition.
Table~\ref{tab:mention-type-distribution} shows the distribution of utterances between mention types, which vary in the explicitness of the signals.
Most utterances carry no explicit addressee cue: implicit and no-reference utterances make up 71\% of MPDD and 79\% of Bazinga!. Role-based references are far more common in MPDD (14.0\%) than in Bazinga! (2.5\%, only 65 utterances).

\begin{table}[h]
\centering
\small
\begin{tabular}{lccccc}
\toprule
\textbf{Dataset} & \textbf{Name} & \textbf{Role} & \textbf{Implicit} & \textbf{NoAd} & \textbf{NoRe} \\
\midrule
MPDD     & 1005 & 1253 & 3390 & 354 & 2976 \\
Bazinga! & 311  & 65   & 972  & 159 & 1080 \\
\bottomrule
\end{tabular}
\caption{Distribution of utterances by addressee mention type in the evaluation split. NoAd means non-addressee and NoRe means no-reference.}
\label{tab:mention-type-distribution}
\end{table}

\paragraph{Results.}

On \textbf{MPDD}, \textbf{name-based} utterances are the easiest and \textbf{non-addressee reference} the hardest for every model, since the latter contains an explicit but misleading person reference. \textsc{Relation} yields large gains on \textbf{role-based} utterances for GPT, LLaMA, and GPT-OSS (+0.07 to +0.08), suggesting that relationship cues help most when the addressee is referred to through a social role; Qwen, already the strongest model on role-based utterances without social information, gains little. Non-addressee utterances also benefit (+0.04 to +0.07 for the same three models), suggesting that knowing who relates to whom helps models look past the misleading mention. GPT improves on every type, including name-based utterances (+0.10), so for this model social information acts as a general disambiguation signal.

On \textbf{Bazinga!}, non-addressee reference is again the hardest type, but the gains are spread more evenly across types, including implicit utterances (+0.03 to +0.08 for Qwen, LLaMA, and GPT-OSS). They are largest for LLaMA (up to +0.11 on no-reference utterances), while GPT barely changes. Role-based results vary across models, as expected given only 65 such utterances.

Overall, social information helps most when the addressee must be inferred from social structure or distinguished from a misleading mention, and the size of its effect depends strongly on the model.

\section{Conclusion}

In this work, we investigated how social relationship information
affects multi-party dialogue agents through a three-stage diagnostic
framework, including addressee recognition, turn-taking judgement,
and response generation.
Across two datasets and four LLMs, socially grounded prompting
is generally beneficial, although its effects vary across stages,
models, and datasets.

For \textbf{addressee recognition}, social information consistently improves performance on MPDD and remains helpful for most models on Bazinga!, confirming that interpersonal cues provide useful signals for resolving whom an utterance is directed to. Our mention-type analysis further shows that these cues are helpful for \textbf{role-based} references, and also help with misleading non-addressee references, although the latter remain the hardest type.

For \textbf{turn-taking judgement}, social information can improve
discrimination by increasing acceptance of gold speakers, reducing
acceptance of non-gold speakers, or both.
However, some models instead become more permissive or conservative.
In \textbf{response generation}, social information generally
yields modest BERTScore gains, while fine-grained relationship
information more consistently improves judge-rated response quality,
particularly in role consistency.

Overall, our findings suggest that \textbf{socially grounded reasoning} 
is an important component of natural multi-party dialogue behavior.
Rather than treating addressee recognition as an isolated task, 
this work highlights the value of studying how social cues influence 
different stages of dialogue participation.
Our results also emphasize that the usefulness of social information 
depends not only on its availability but also on its granularity 
and how models utilize it.
Understanding whom to respond to, when to speak, and what to say 
requires more than interpreting the linguistic content of a conversation.
We hope this setting can serve as a useful framework for future 
research on socially aware conversational agents.

\section*{Limitations}

Both MPDD and the reconstructed \textit{Bazinga!} subset are derived from scripted TV dialogue, which may not fully capture the spontaneity, overlapping speech, and disfluencies of real-world multi-party interactions, such as everyday family or workplace conversations.

In addition, the \textit{Bazinga!} subset is relatively small and relies partly on manually inferred relations, which may introduce noise and weaken the reliability of cross-lingual comparisons. The datasets are also text-only, excluding multimodal cues such as gaze, intonation, and gesture that are often important for addressee resolution. Moreover, the turn-taking and response-generation stages should be interpreted as diagnostic probes rather than complete formulations of real-world conversational behavior. Finally, our fine-tuning experiments (Appendix~\ref{app:finetune}) cover only addressee
recognition with two open-weight LLMs; turn-taking judgement and response
generation are evaluated through prompting only.
Moreover, although folds are split at the dialogue level (as contiguous blocks of
dialogues, which reduces but does not remove overlap), the same characters recur
across folds, so fine-tuned models may partly learn character-specific addressing
patterns rather than general social reasoning.

\section*{Acknowledgments}

This work was supported by the National Science and Technology Council (NSTC), Taiwan, under Grant Nos. NSTC 114-2221-E-002-070-MY3 and NSTC 115-2634-F-002-012. Additional financial support was provided by the Featured Area Research Center Program within the framework of the Higher Education Sprout Project of the Ministry of Education, Taiwan (Grant No. 115L900901).

\bibliography{eacl_bib}

@inproceedings{chen2020mpdd,
  title={MPDD: A Multi-Party Dialogue Dataset for Analysis of Emotions and Interpersonal Relationships},
  author={Chen, Yi-Ting and Huang, Hen-Hsen and Chen, Hsin-Hsi},
  booktitle={Proceedings of the Twelfth Language Resources and Evaluation Conference},
  pages={610--614},
  year={2020},
  address={Marseille, France},
  publisher={European Language Resources Association}
}

@inproceedings{madotto2022bazinga,
  title={Bazinga! A Dataset for Multi-Party Dialogues Structuring},
  author={Lerner, Paul and Bergo{\"e}nd, Juliette and Guinaudeau, Camille and Bredin, Herv{\'e} and Maurice, Benjamin and Lefevre, Sharleyne and Bouteiller, Martin and Berhe, Aman and Galmant, L{\'e}o and Yin, Ruiqing and Barras, Claude},
  booktitle={Proceedings of the Thirteenth Language Resources and Evaluation Conference},
  pages={3434--3441},
  year={2022},
  address={Marseille, France},
  publisher={European Language Resources Association}
}

@inproceedings{ouchi2016addressee,
  title={Addressee and Response Selection for Multi-Party Conversation},
  author={Ouchi, Hiroki and Tsuboi, Yuta},
  booktitle={Proceedings of the 2016 Conference on Empirical Methods in Natural Language Processing},
  pages={2133--2143},
  year={2016},
  address={Austin, Texas},
  publisher={Association for Computational Linguistics}
}

@inproceedings{zhang2018sirnn,
  title={Addressee and Response Selection in Multi-Party Conversations with Speaker Interaction RNNs},
  author={Zhang, Rui and Lee, Honglak and Polymenakos, Lazaros and Radev, Dragomir R.},
  booktitle={Proceedings of the Thirty-Second AAAI Conference on Artificial Intelligence},
  pages={5690--5697},
  year={2018},
  publisher={AAAI Press}
}

@inproceedings{le2019addressee,
  title={Who Is Speaking to Whom? Learning to Identify Utterance Addressee in Multi-Party Conversations},
  author={Le, Ran and Hu, Wenpeng and Shang, Mingyue and You, Zhenjun and Bing, Lidong and Zhao, Dongyan and Yan, Rui},
  booktitle={Proceedings of the 2019 Conference on Empirical Methods in Natural Language Processing and the 9th International Joint Conference on Natural Language Processing (EMNLP-IJCNLP)},
  pages={1909--1919},
  year={2019},
  address={Hong Kong, China},
  publisher={Association for Computational Linguistics}
}

@inproceedings{gu2021mpcbert,
  title={MPC-BERT: A Pre-Trained Language Model for Multi-Party Conversation Understanding},
  author={Gu, Jia-Chen and Tao, Chongyang and Ling, Zhen-Hua and Xu, Can and Geng, Xiubo and Jiang, Daxin},
  booktitle={Proceedings of the 59th Annual Meeting of the Association for Computational Linguistics and the 11th International Joint Conference on Natural Language Processing (Volume 1: Long Papers)},
  pages={3682--3692},
  year={2021},
  address={Online},
  publisher={Association for Computational Linguistics}
}

@inproceedings{gu2023gift,
  title={GIFT: Graph-Induced Fine-Tuning for Multi-Party Conversation Understanding},
  author={Gu, Jia-Chen and Ling, Zhenhua and Liu, Quan and Liu, Cong and Hu, Guoping},
  booktitle={Proceedings of the 61st Annual Meeting of the Association for Computational Linguistics (Volume 1: Long Papers)},
  pages={11645--11658},
  year={2023},
  address={Toronto, Canada},
  publisher={Association for Computational Linguistics}
}

@inproceedings{wang2024replytostructure,
  title={Improving Dialogue Discourse Parsing via Reply-to Structures of Addressee Recognition},
  author={Fan, Yaxin and Jiang, Feng and Li, Peifeng and Kong, Fang and Zhu, Qiaoming},
  booktitle={Proceedings of the 2023 Conference on Empirical Methods in Natural Language Processing},
  pages={8484--8495},
  year={2023},
  address={Singapore},
  publisher={Association for Computational Linguistics}
}

@inproceedings{mahajan2024persona,
  title={Persona-aware Multi-party Conversation Response Generation},
  author={Mahajan, Khyati and Shaikh, Samira},
  booktitle={Proceedings of the 2024 Joint International Conference on Computational Linguistics, Language Resources and Evaluation (LREC-COLING 2024)},
  pages={12712--12723},
  year={2024},
  address={Torino, Italia},
  publisher={ELRA and ICCL}
}

@article{lee2025mirror,
  title={MIRROR: Multi-party Dialogue Generation Based on Interpersonal Relationship-Aware Persona Retrieval},
  author={Lee, Seungmi and Shim, Midan and Lee, Kyong-Ho},
  journal={Expert Systems with Applications},
  volume={276},
  pages={127141},
  year={2025},
  doi={10.1016/j.eswa.2025.127141}
}

@inproceedings{tan2023evaluations,
  title={Is ChatGPT a Good Multi-Party Conversation Solver?},
  author={Tan, Chao-Hong and Gu, Jia-Chen and Ling, Zhen-Hua},
  booktitle={Findings of the Association for Computational Linguistics: EMNLP 2023},
  pages={4905--4915},
  year={2023},
  address={Singapore},
  publisher={Association for Computational Linguistics}
}

@inproceedings{penzo2024hangover,
  title={Do LLMs Suffer from Multi-Party Hangover? A Diagnostic Approach to Addressee Recognition and Response Selection in Conversations},
  author={Penzo, Nicol{\`o} and Sajedinia, Maryam and Lepri, Bruno and Tonelli, Sara and Guerini, Marco},
  booktitle={Proceedings of the 2024 Conference on Empirical Methods in Natural Language Processing},
  pages={11210--11233},
  year={2024},
  address={Miami, Florida, USA},
  publisher={Association for Computational Linguistics}
}

@inproceedings{inoue2025benchmark,
  title={An LLM Benchmark for Addressee Recognition in Multi-modal Multi-party Dialogue},
  author={Inoue, Koji and Lala, Divesh and Elmers, Mikey and Ochi, Keiko and Kawahara, Tatsuya},
  booktitle={Proceedings of the 15th International Workshop on Spoken Dialogue Systems Technology},
  pages={330--334},
  year={2025},
  address={Bilbao, Spain},
  publisher={Association for Computational Linguistics}
}

@inproceedings{li2023empretrain,
  title={EM Pre-training for Multi-party Dialogue Response Generation},
  author={Li, Yiyang and Zhao, Hai},
  booktitle={Proceedings of the 61st Annual Meeting of the Association for Computational Linguistics (Volume 1: Long Papers)},
  pages={92--103},
  year={2023},
  address={Toronto, Canada},
  publisher={Association for Computational Linguistics}
}

@inproceedings{fan2024rltrc,
  title={Improving Multi-party Dialogue Generation via Topic and Rhetorical Coherence},
  author={Fan, Yaxin and Li, Peifeng and Zhu, Qiaoming},
  booktitle={Proceedings of the 2024 Conference on Empirical Methods in Natural Language Processing},
  pages={3240--3253},
  year={2024},
  address={Miami, Florida, USA},
  publisher={Association for Computational Linguistics}
}

@misc{wang2024mupas,
  title={Multi-Party Supervised Fine-tuning of Language Models for Multi-Party Dialogue Generation},
  author={Wang, Xiaoyu and Xi, Ningyuan and Chen, Teng and Gu, Qingqing and Zhao, Yue and Chen, Xiaokai and Jiang, Zhonglin and Chen, Yong and Luo, Ji},
  year={2024},
  eprint={2412.05342},
  archivePrefix={arXiv},
  primaryClass={cs.CL}
}

@misc{openai2025gpt5mini,
  author       = {OpenAI},
  title        = {GPT-5 mini},
  year         = {2025},
  howpublished = {\url{https://developers.openai.com/api/docs/models/gpt-5-mini}},
  note         = {OpenAI API model documentation. Accessed: 2026-04-01}
}

@article{yang2025qwen3,
  title   = {Qwen3 Technical Report},
  author={An Yang and Anfeng Li and Baosong Yang and Beichen Zhang and Binyuan Hui and Bo Zheng and Bowen Yu and Chang Gao and Chengen Huang and Chenxu Lv and Chujie Zheng and Dayiheng Liu and Fan Zhou and Fei Huang and Feng Hu and Hao Ge and Haoran Wei and Huan Lin and Jialong Tang and Jian Yang and Jianhong Tu and Jianwei Zhang and Jianxin Yang and Jiaxi Yang and Jing Zhou and Jingren Zhou and Junyang Lin and Kai Dang and Keqin Bao and Kexin Yang and Le Yu and Lianghao Deng and Mei Li and Mingfeng Xue and Mingze Li and Pei Zhang and Peng Wang and Qin Zhu and Rui Men and Ruize Gao and Shixuan Liu and Shuang Luo and Tianhao Li and Tianyi Tang and Wenbiao Yin and Xingzhang Ren and Xinyu Wang and Xinyu Zhang and Xuancheng Ren and Yang Fan and Yang Su and Yichang Zhang and Yinger Zhang and Yu Wan and Yuqiong Liu and Zekun Wang and Zeyu Cui and Zhenru Zhang and Zhipeng Zhou and Zihan Qiu},
  journal = {arXiv preprint arXiv:2505.09388},
  year    = {2025}
}

@misc{dubey2024llama3,
  title         = {The Llama 3 Herd of Models},
  author        = {Aaron Grattafiori and Abhimanyu Dubey and Abhinav Jauhri and Abhinav Pandey and Abhishek Kadian and Ahmad Al-Dahle and Aiesha Letman and Akhil Mathur and Alan Schelten and Alex Vaughan and Amy Yang and Angela Fan and Anirudh Goyal and Anthony Hartshorn and Aobo Yang and Archi Mitra and Archie Sravankumar and Artem Korenev and Arthur Hinsvark and Arun Rao and Aston Zhang and Aurelien Rodriguez and Austen Gregerson and Ava Spataru and Baptiste Roziere and Bethany Biron and Binh Tang and Bobbie Chern and Charlotte Caucheteux and Chaya Nayak and Chloe Bi and Chris Marra and Chris McConnell and Christian Keller and Christophe Touret and Chunyang Wu and Corinne Wong and Cristian Canton Ferrer and Cyrus Nikolaidis and Damien Allonsius and Daniel Song and Danielle Pintz and Danny Livshits and Danny Wyatt and David Esiobu and Dhruv Choudhary and Dhruv Mahajan and Diego Garcia-Olano and Diego Perino and Dieuwke Hupkes and Egor Lakomkin and Ehab AlBadawy and Elina Lobanova and Emily Dinan and Eric Michael Smith and Filip Radenovic and Francisco Guzmán and Frank Zhang and Gabriel Synnaeve and Gabrielle Lee and Georgia Lewis Anderson and Govind Thattai and Graeme Nail and Gregoire Mialon and Guan Pang and Guillem Cucurell and Hailey Nguyen and Hannah Korevaar and Hu Xu and Hugo Touvron and Iliyan Zarov and Imanol Arrieta Ibarra and Isabel Kloumann and Ishan Misra and Ivan Evtimov and Jack Zhang and Jade Copet and Jaewon Lee and Jan Geffert and Jana Vranes and Jason Park and Jay Mahadeokar and Jeet Shah and Jelmer van der Linde and Jennifer Billock and Jenny Hong and Jenya Lee and Jeremy Fu and Jianfeng Chi and Jianyu Huang and Jiawen Liu and Jie Wang and Jiecao Yu and Joanna Bitton and Joe Spisak and Jongsoo Park and Joseph Rocca and Joshua Johnstun and Joshua Saxe and Junteng Jia and Kalyan Vasuden Alwala and Karthik Prasad and Kartikeya Upasani and Kate Plawiak and Ke Li and Kenneth Heafield and Kevin Stone and Khalid El-Arini and Krithika Iyer and Kshitiz Malik and Kuenley Chiu and Kunal Bhalla and Kushal Lakhotia and Lauren Rantala-Yeary and Laurens van der Maaten and Lawrence Chen and Liang Tan and Liz Jenkins and Louis Martin and Lovish Madaan and Lubo Malo and Lukas Blecher and Lukas Landzaat and Luke de Oliveira and Madeline Muzzi and Mahesh Pasupuleti and Mannat Singh and Manohar Paluri and Marcin Kardas and Maria Tsimpoukelli and Mathew Oldham and Mathieu Rita and Maya Pavlova and Melanie Kambadur and Mike Lewis and Min Si and Mitesh Kumar Singh and Mona Hassan and Naman Goyal and Narjes Torabi and Nikolay Bashlykov and Nikolay Bogoychev and Niladri Chatterji and Ning Zhang and Olivier Duchenne and Onur Çelebi and Patrick Alrassy and Pengchuan Zhang and Pengwei Li and Petar Vasic and Peter Weng and Prajjwal Bhargava and Pratik Dubal and Praveen Krishnan and Punit Singh Koura and Puxin Xu and Qing He and Qingxiao Dong and Ragavan Srinivasan and Raj Ganapathy and Ramon Calderer and Ricardo Silveira Cabral and Robert Stojnic and Roberta Raileanu and Rohan Maheswari and Rohit Girdhar and Rohit Patel and Romain Sauvestre and Ronnie Polidoro and Roshan Sumbaly and Ross Taylor and Ruan Silva and Rui Hou and Rui Wang and Saghar Hosseini and Sahana Chennabasappa and Sanjay Singh and Sean Bell and Seohyun Sonia Kim and Sergey Edunov and Shaoliang Nie and Sharan Narang and Sharath Raparthy and Sheng Shen and Shengye Wan and Shruti Bhosale and Shun Zhang and Simon Vandenhende and Soumya Batra and Spencer Whitman and Sten Sootla and Stephane Collot and Suchin Gururangan and Sydney Borodinsky and Tamar Herman and Tara Fowler and Tarek Sheasha and Thomas Georgiou and Thomas Scialom and Tobias Speckbacher and Todor Mihaylov and Tong Xiao and Ujjwal Karn and Vedanuj Goswami and Vibhor Gupta and Vignesh Ramanathan and Viktor Kerkez and Vincent Gonguet and Virginie Do and Vish Vogeti and Vítor Albiero and Vladan Petrovic and Weiwei Chu and Wenhan Xiong and Wenyin Fu and Whitney Meers and Xavier Martinet and Xiaodong Wang and Xiaofang Wang and Xiaoqing Ellen Tan and Xide Xia and Xinfeng Xie and Xuchao Jia and Xuewei Wang and Yaelle Goldschlag and Yashesh Gaur and Yasmine Babaei and Yi Wen and Yiwen Song and Yuchen Zhang and Yue Li and Yuning Mao and Zacharie Delpierre Coudert and Zheng Yan and Zhengxing Chen and Zoe Papakipos and Aaditya Singh and Aayushi Srivastava and Abha Jain and Adam Kelsey and Adam Shajnfeld and Adithya Gangidi and Adolfo Victoria and Ahuva Goldstand and Ajay Menon and Ajay Sharma and Alex Boesenberg and Alexei Baevski and Allie Feinstein and Amanda Kallet and Amit Sangani and Amos Teo and Anam Yunus and Andrei Lupu and Andres Alvarado and Andrew Caples and Andrew Gu and Andrew Ho and Andrew Poulton and Andrew Ryan and Ankit Ramchandani and Annie Dong and Annie Franco and Anuj Goyal and Aparajita Saraf and Arkabandhu Chowdhury and Ashley Gabriel and Ashwin Bharambe and Assaf Eisenman and Azadeh Yazdan and Beau James and Ben Maurer and Benjamin Leonhardi and Bernie Huang and Beth Loyd and Beto De Paola and Bhargavi Paranjape and Bing Liu and Bo Wu and Boyu Ni and Braden Hancock and Bram Wasti and Brandon Spence and Brani Stojkovic and Brian Gamido and Britt Montalvo and Carl Parker and Carly Burton and Catalina Mejia and Ce Liu and Changhan Wang and Changkyu Kim and Chao Zhou and Chester Hu and Ching-Hsiang Chu and Chris Cai and Chris Tindal and Christoph Feichtenhofer and Cynthia Gao and Damon Civin and Dana Beaty and Daniel Kreymer and Daniel Li and David Adkins and David Xu and Davide Testuggine and Delia David and Devi Parikh and Diana Liskovich and Didem Foss and Dingkang Wang and Duc Le and Dustin Holland and Edward Dowling and Eissa Jamil and Elaine Montgomery and Eleonora Presani and Emily Hahn and Emily Wood and Eric-Tuan Le and Erik Brinkman and Esteban Arcaute and Evan Dunbar and Evan Smothers and Fei Sun and Felix Kreuk and Feng Tian and Filippos Kokkinos and Firat Ozgenel and Francesco Caggioni and Frank Kanayet and Frank Seide and Gabriela Medina Florez and Gabriella Schwarz and Gada Badeer and Georgia Swee and Gil Halpern and Grant Herman and Grigory Sizov and Guangyi and Zhang and Guna Lakshminarayanan and Hakan Inan and Hamid Shojanazeri and Han Zou and Hannah Wang and Hanwen Zha and Haroun Habeeb and Harrison Rudolph and Helen Suk and Henry Aspegren and Hunter Goldman and Hongyuan Zhan and Ibrahim Damlaj and Igor Molybog and Igor Tufanov and Ilias Leontiadis and Irina-Elena Veliche and Itai Gat and Jake Weissman and James Geboski and James Kohli and Janice Lam and Japhet Asher and Jean-Baptiste Gaya and Jeff Marcus and Jeff Tang and Jennifer Chan and Jenny Zhen and Jeremy Reizenstein and Jeremy Teboul and Jessica Zhong and Jian Jin and Jingyi Yang and Joe Cummings and Jon Carvill and Jon Shepard and Jonathan McPhie and Jonathan Torres and Josh Ginsburg and Junjie Wang and Kai Wu and Kam Hou U and Karan Saxena and Kartikay Khandelwal and Katayoun Zand and Kathy Matosich and Kaushik Veeraraghavan and Kelly Michelena and Keqian Li and Kiran Jagadeesh and Kun Huang and Kunal Chawla and Kyle Huang and Lailin Chen and Lakshya Garg and Lavender A and Leandro Silva and Lee Bell and Lei Zhang and Liangpeng Guo and Licheng Yu and Liron Moshkovich and Luca Wehrstedt and Madian Khabsa and Manav Avalani and Manish Bhatt and Martynas Mankus and Matan Hasson and Matthew Lennie and Matthias Reso and Maxim Groshev and Maxim Naumov and Maya Lathi and Meghan Keneally and Miao Liu and Michael L. Seltzer and Michal Valko and Michelle Restrepo and Mihir Patel and Mik Vyatskov and Mikayel Samvelyan and Mike Clark and Mike Macey and Mike Wang and Miquel Jubert Hermoso and Mo Metanat and Mohammad Rastegari and Munish Bansal and Nandhini Santhanam and Natascha Parks and Natasha White and Navyata Bawa and Nayan Singhal and Nick Egebo and Nicolas Usunier and Nikhil Mehta and Nikolay Pavlovich Laptev and Ning Dong and Norman Cheng and Oleg Chernoguz and Olivia Hart and Omkar Salpekar and Ozlem Kalinli and Parkin Kent and Parth Parekh and Paul Saab and Pavan Balaji and Pedro Rittner and Philip Bontrager and Pierre Roux and Piotr Dollar and Polina Zvyagina and Prashant Ratanchandani and Pritish Yuvraj and Qian Liang and Rachad Alao and Rachel Rodriguez and Rafi Ayub and Raghotham Murthy and Raghu Nayani and Rahul Mitra and Rangaprabhu Parthasarathy and Raymond Li and Rebekkah Hogan and Robin Battey and Rocky Wang and Russ Howes and Ruty Rinott and Sachin Mehta and Sachin Siby and Sai Jayesh Bondu and Samyak Datta and Sara Chugh and Sara Hunt and Sargun Dhillon and Sasha Sidorov and Satadru Pan and Saurabh Mahajan and Saurabh Verma and Seiji Yamamoto and Sharadh Ramaswamy and Shaun Lindsay and Shaun Lindsay and Sheng Feng and Shenghao Lin and Shengxin Cindy Zha and Shishir Patil and Shiva Shankar and Shuqiang Zhang and Shuqiang Zhang and Sinong Wang and Sneha Agarwal and Soji Sajuyigbe and Soumith Chintala and Stephanie Max and Stephen Chen and Steve Kehoe and Steve Satterfield and Sudarshan Govindaprasad and Sumit Gupta and Summer Deng and Sungmin Cho and Sunny Virk and Suraj Subramanian and Sy Choudhury and Sydney Goldman and Tal Remez and Tamar Glaser and Tamara Best and Thilo Koehler and Thomas Robinson and Tianhe Li and Tianjun Zhang and Tim Matthews and Timothy Chou and Tzook Shaked and Varun Vontimitta and Victoria Ajayi and Victoria Montanez and Vijai Mohan and Vinay Satish Kumar and Vishal Mangla and Vlad Ionescu and Vlad Poenaru and Vlad Tiberiu Mihailescu and Vladimir Ivanov and Wei Li and Wenchen Wang and Wenwen Jiang and Wes Bouaziz and Will Constable and Xiaocheng Tang and Xiaojian Wu and Xiaolan Wang and Xilun Wu and Xinbo Gao and Yaniv Kleinman and Yanjun Chen and Ye Hu and Ye Jia and Ye Qi and Yenda Li and Yilin Zhang and Ying Zhang and Yossi Adi and Youngjin Nam and Yu and Wang and Yu Zhao and Yuchen Hao and Yundi Qian and Yunlu Li and Yuzi He and Zach Rait and Zachary DeVito and Zef Rosnbrick and Zhaoduo Wen and Zhenyu Yang and Zhiwei Zhao and Zhiyu Ma},
  year          = {2024},
  eprint        = {2407.21783},
  archivePrefix = {arXiv},
  primaryClass  = {cs.AI},
  url           = {https://arxiv.org/abs/2407.21783}
}

@misc{openai2025gptoss120b,
  author       = {OpenAI},
  title        = {gpt-oss-120b},
  year         = {2025},
  howpublished = {\url{https://developers.openai.com/api/docs/models/gpt-oss-120b}},
  note         = {OpenAI API model documentation. Accessed: 2026-04-01}
}

@inproceedings{jiang-etal-2022-crecil,
    title = "The {CRECIL} Corpus: a New Dataset for Extraction of Relations between Characters in {C}hinese Multi-party Dialogues",
    author = "Jiang, Yuru and Xu, Yang and Zhan, Yuhang and He, Weikai and Wang, Yilin and Xi, Zixuan and Wang, Meiyun and Li, Xinyu and Li, Yu and Yu, Yanchao",
    booktitle = "Proceedings of the Thirteenth Language Resources and Evaluation Conference",
    month = jun,
    year = "2022",
    address = "Marseille, France",
    publisher = "European Language Resources Association",
    url = "https://aclanthology.org/2022.lrec-1.250/",
    pages = "2337--2344"
}

@misc{yan-etal-2024-identifying,
    title = "Identifying Speakers and Addressees of Quotations in Novels with Prompt Learning",
    author = "Yan, Yuchen and Zhao, Hanjie and Zhu, Senbin and Liu, Hongde and Zhang, Zhihong and Jia, Yuxiang",
    year = "2024",
    eprint = "2408.09452",
    archivePrefix = "arXiv",
    primaryClass = "cs.CL",
    url = "https://arxiv.org/abs/2408.09452"
}

@inproceedings{yu-etal-2020-dialogue,
    title = "Dialogue-Based Relation Extraction",
    author = "Yu, Dian and Sun, Kai and Cardie, Claire and Yu, Dong",
    booktitle = "Proceedings of the 58th Annual Meeting of the Association for Computational Linguistics",
    month = jul,
    year = "2020",
    address = "Online",
    publisher = "Association for Computational Linguistics",
    url = "https://aclanthology.org/2020.acl-main.444/",
    doi = "10.18653/v1/2020.acl-main.444",
    pages = "4927--4940"
}

@inproceedings{papay-pado-2020-riqua,
    title = "{R}i{Q}u{A}: A Corpus of Rich Quotation Annotation for {E}nglish Literary Text",
    author = "Papay, Sean and Pad{\'o}, Sebastian",
    booktitle = "Proceedings of the Twelfth Language Resources and Evaluation Conference",
    month = may,
    year = "2020",
    address = "Marseille, France",
    publisher = "European Language Resources Association",
    url = "https://aclanthology.org/2020.lrec-1.104/",
    pages = "835--841"
}

@misc{meta2024llama33,
    author = {{Meta AI}},
    title  = {{Llama 3.3 70B Instruct Model Card}},
    year   = {2024},
    url    = {https://huggingface.co/meta-llama/Llama-3.3-70B-Instruct}
}

\appendix

\section{Mapping from Relation Types to Position Categories}\label{sec:mapping}
\begin{table*}[h]
\centering
\begin{tabular}{p{1.1cm} p{11cm}}
\toprule
\textbf{Position} & \textbf{Relation Types} \\
\midrule
Superior & parent, parent-in-law, grandparent, teacher, boss, other superior \\ \midrule
Peer     & spouse, brothers and sisters, other peer, classmate, colleague, partner, couple, friend, enemy, consignee, consignor, stranger, unknown \\ \midrule
Inferior & child, son/daughter-in-law, other inferior, grandchild, student, subordinate \\
\bottomrule
\end{tabular}
\caption{Mapping from fine-grained social relations to position categories.}
\label{tab:relation-map}
\end{table*}

To simplify the use of social relation information in our prompts, we group the fine-grained relation types annotated in the MPDD dataset into three abstract position categories: \textit{superior}, \textit{peer}, and \textit{inferior}.
This abstraction allows the model to access relational cues without relying on specific labels. 
The full mapping used in our experiments is shown in Table~\ref{tab:relation-map}.

\section{Reconstructing the Bazinga! Corpus}
\label{tab:bazinga_construct}

This appendix describes how we reconstruct an English multi-party dialogue benchmark (\textit{Bazinga!}) in an MPDD-compatible format with addressee relations.

\subsection{Relationship Table Construction}
We first curate a show-specific relationship table using the same relation ontology as MPDD.
Concretely, given MPDD's predefined relation labels (e.g., \texttt{parent}, \texttt{spouse}, \texttt{boss}, \texttt{friend}), we search external resources (e.g., character lists, fandom wikis, episode guides) to verify whether each MPDD relation type is instantiated in each drama.
We then record directed relations as a mapping \texttt{rel\_map[speaker][addressee] = relation}.
For asymmetric relations, we additionally define an inverse mapping (e.g., \texttt{parent}$\leftrightarrow$\texttt{child}, \texttt{boss}$\leftrightarrow$\texttt{subordinate}, \texttt{teacher}$\leftrightarrow$\texttt{student}) to allow lookup in either direction.

\subsection{Turn Cleaning and Relation Attachment}
We convert raw subtitle turns into MPDD-style turns (\texttt{speaker}, \texttt{utterance}, \texttt{listener}).
Each turn is filtered by the following rules:
(1) drop turns with invalid speakers (e.g., \texttt{unknown}, \texttt{group}, \texttt{narrator});
(2) remove non-linguistic stage directions (e.g., bracketed actions, sound effects, punctuation-only lines);
(3) clean addressees by removing placeholders (e.g., \texttt{everyone}, \texttt{all});
(4) attach relations by keeping only addressees that have a known relation with the speaker under \texttt{rel\_map}, using inverse lookup when necessary.
If no addressee remains after relation attachment, the turn is discarded.

\subsection{Dialogue Segmentation and Filtering}
We segment dialogues using a boundary rule: any discarded turn triggers a cut, splitting the original sequence into maximal contiguous chunks of valid turns.
We then apply the following MPDD-style constraints to each chunk (or window):
(1) length between 4 and 15 turns;
(2) number of unique speakers between 3 and 7;
(3) closure constraint: every addressee appearing in the chunk must also appear as a speaker within the same chunk.
For sequences longer than 15 turns, we apply sliding windows with size 15 and step 13, and retain only windows satisfying the above constraints.

\subsection{Quality Control and Reporting}
Table~\ref{tab:bazinga_relation_dist} shows the resulting relation distribution. During construction, we also inspected the most frequent speaker--addressee pairs without a relationship entry in \texttt{rel\_map} to identify missing relations.

\begin{table*}[h]
\centering
\setlength{\tabcolsep}{6pt}
\begin{tabular}{llc|llc}
\toprule
\textbf{Position} & \textbf{Relation} & \textbf{\% of All} & \textbf{Position} & \textbf{Relation} & \textbf{\% of All} \\
\midrule
Superior & parent           & 4.24\%  & Peer & spouse               & 4.75\% \\
         & teacher          & 7.59\%  &      & brothers and sisters & 3.45\% \\
         & boss             & 5.65\%  &      & other peer           & 3.38\% \\
         & other superior   & 0.58\%  &      & classmate            & 2.98\% \\
\cmidrule(r){1-3}
Inferior & child            & 4.10\%  &      & colleague            & 5.61\% \\
         & student          & 6.76\%  &      & partner              & 3.92\% \\
         & subordinate      & 6.11\%  &      & couple               & 13.70\% \\
         & other inferior   & 1.12\%  &      & friend               & 22.47\% \\
         &                  &         &      & enemy                & 3.60\% \\
\bottomrule
\end{tabular}
\caption{
Relation type distribution in the reconstructed Bazinga! corpus.
Percentages are computed over all relations, not within each position.
}
\label{tab:bazinga_relation_dist}
\end{table*}

\section{Implementation Notes}

We release the full codebase for reproducing all experiments at~\url{https://github.com/jingjin412/KYR-TYT/}. Here we report only the key inference settings that directly affect experimental outcomes. For addressee recognition and turn-taking judgement, we use deterministic decoding (temperature 0 when supported, or greedy decoding for local HuggingFace inference). The maximum generation length is set per model backend. Local open-weight models are run with 4-bit quantization when applicable. Full implementation details, prompt files, parsing rules, and dataset processing scripts are provided in the released repository.

\section{Prompt Templates for the Three-Stage Framework}
\label{app:prompts}

Across all three stages, we use task-specific prompt templates with a shared mechanism for injecting participant-level social information. In each prompt, the placeholder \texttt{<relation\_list>} is used to control the social-information condition: it is left empty under \textsc{None}, replaced with position-level metadata under \textsc{Position}, and replaced with fine-grained interpersonal relations under \textsc{Relation}. All other parts of the prompt remain unchanged within each stage.

\noindent\textbf{Example of inserted social-information block.}

\begin{quote}\small
\textbf{Position block}\\
Liu's mom is Liu's dad's Peer\\
Xiaobao is Liu's mom's Inferior\\
Xiaobao is Liu's dad's Inferior

\vspace{0.5em}
\textbf{Relation block}\\
Liu's mom is Liu's dad's Spouse\\
Xiaobao is Liu's mom's Child\\
Xiaobao is Liu's dad's Child
\end{quote}

\subsection{Addressee Recognition Prompt Template}

For the main addressee recognition experiments, we use a manually written prompt template with a fixed instruction and output format. The prompt provides the participant list, the dialogue context, and the target utterance, and asks the model to identify the intended addressee(s) from the provided character list.

\noindent\textbf{Prompt template.}

\begin{quote}\small
Your task is to determine the listener in a conversation. You will be given a dialogue segment and character information, and you need to identify who \texttt{<person>} is speaking to when they say "\texttt{<utterance>}".

\textbf{\#\#\# Character Information \#\#\#}\\
Character list: \texttt{<all\_char>}\\
\texttt{<relation\_list>}

\textbf{\#\#\# Conversation Content \#\#\#}\\
\texttt{<dialogue>}

\textbf{\#\#\# Task \#\#\#}\\
Please determine the target(s) of the sentence "\texttt{<utterance>}" spoken by \texttt{<person>}, and select the appropriate character(s) from the character list.\\
- If the target may be multiple people, select all appropriate characters.\\
- Only choose characters that appear in the character list; do not include characters not listed.\\
- Provide the answer directly without explanation.

\textbf{\#\#\# Answer Format \#\#\#}\\
- If the targets are "Amber and Justin," answer: "Answer: Amber,Justin"\\
- If the target is only "Amber," answer: "Answer: Amber"\\
- Reply the \texttt{FULL} name of the character in character list; do not use abbreviation or short name.
\end{quote}

\subsection{Turn-Taking Judgement Prompt Template}

For the turn-taking judgement stage, we assign the model the identity of each participant and ask whether it is plausible that this character would speak next given the preceding dialogue context. Concretely, the placeholder \texttt{<agent\_name>} in the prompt is instantiated with each participant for that dialogue instance. The model is then asked to make a binary plausibility judgement, rather than to select the next speaker from all participants.

\noindent\textbf{Prompt template.}

\begin{quote}\small
You are now playing the role of "\texttt{<agent\_name>}". Please determine whether it is possible that the next sentence will be your turn to speak based on the following dialogue.

\textbf{\#\#\# Character and Relationship \#\#\#}\\
\texttt{<all\_char>}
\texttt{<relation\_list>}

\textbf{\#\#\# Dialogue Context \#\#\#}\\
\texttt{<dialogue>}

\texttt{<last\_speaker>} said: "\texttt{<utterance>}"

Please determine whether it is possible that you will be the next speaker. Answer only "yes" or "no".\\
Format: Answer: yes or Answer: no
\end{quote}

\subsection{Response Generation Prompt Template}

For the response generation stage, we again assign the model the identity of the gold next speaker. In the prompt, \texttt{<agent\_name>} is instantiated with that speaker, and the model is asked to generate the next utterance in character based on the preceding dialogue context. As in the other stages, the placeholder \texttt{<relation\_list>} is left empty under \textsc{None}, replaced with position-level metadata under \textsc{Position}, and replaced with fine-grained interpersonal relations under \textsc{Relation}.

\noindent\textbf{Prompt template.}

\begin{quote}\small
You are now playing the role of "\texttt{<agent\_name>}". Please generate a natural and appropriate response in the role of the character, without breaking character.

\textbf{\#\#\# Character and Relationship \#\#\#}\\
\texttt{<relation\_list>}

\textbf{\#\#\# Dialogue Context \#\#\#}\\
\texttt{<dialogue>}

\texttt{<last\_speaker>} said: "\texttt{<utterance>}"

Please generate a response as \texttt{<agent\_name>}.\\
Format: Response: \texttt{<your response>}
\end{quote}

\section{Prompt Sensitivity Analysis for Addressee Recognition}
\label{app: prompt sensitivity}
We additionally test whether the effect of social information is sensitive to prompt wording. Starting from the manually written addressee recognition prompt used in the main experiments, we use an LLM to generate five rewritten variants (P1--P5) that preserve the same task definition and answer constraints but differ in phrasing.

Due to resource and efficiency considerations, we conduct this analysis only on LLaMA-3.1-8B and Qwen3-32B. These two models provide a useful contrast in model behavior while keeping the supplementary evaluation tractable.

Table~\ref{tab:prompt-sensitivity} shows the resulting F1 scores on Bazinga! and MPDD. Absolute performance varies across prompt variants, especially for LLaMA. Nevertheless, adding \textsc{Position} or \textsc{Relation} information is usually beneficial, with mostly non-negative gains and many statistically significant improvements over the \textsc{None} condition. Overall, these results suggest that the benefit of social information is not tied to a single prompt wording. The full text of prompts P1--P5 is provided in the released code repository.

\begin{table*}[h]
\centering
\small
\setlength{\tabcolsep}{6pt}
\caption{Prompt sensitivity results for addressee recognition (F1). P1--P5 denote five prompt rewrites generated from the manually written base prompt. Diff is computed as \textsc{None} $-$ w/ Info. A $\dag$ indicates a statistically significant difference from the \textsc{None} condition under a paired t-test.}
\label{tab:prompt-sensitivity}
\begin{tabular}{llcccccc}
\toprule
Dataset & Model & Prompt & None & Position & Relation & $\Delta_{\text{Pos}}$ & $\Delta_{\text{Rel}}$ \\
\midrule
\multirow{10}{*}{MPDD}
& \multirow{5}{*}{LLaMA}
& P1 & 0.5921 & 0.6020 & 0.5984 & \colordiff{-0.0099}$^\dag$ & \colordiff{-0.0063} \\
& & P2 & 0.6711 & 0.6796 & 0.6720 & \colordiff{-0.0085} & \colordiff{-0.0009} \\
& & P3 & 0.6472 & 0.6748 & 0.6711 & \colordiff{-0.0276}$^\dag$ & \colordiff{-0.0239}$^\dag$ \\
& & P4 & 0.6024 & 0.6015 & 0.6022 & \colordiff{0.0009} & \colordiff{0.0002} \\
& & P5 & 0.7103 & 0.7231 & 0.7393 & \colordiff{-0.0128}$^\dag$ & \colordiff{-0.0290}$^\dag$ \\
\cmidrule(lr){2-8}
& \multirow{5}{*}{Qwen}
& P1 & 0.8017 & 0.8084 & 0.8196 & \colordiff{-0.0067} & \colordiff{-0.0179}$^\dag$ \\
& & P2 & 0.8044 & 0.8188 & 0.8187 & \colordiff{-0.0144}$^\dag$ & \colordiff{-0.0143}$^\dag$ \\
& & P3 & 0.7932 & 0.8123 & 0.8142 & \colordiff{-0.0191}$^\dag$ & \colordiff{-0.0210}$^\dag$ \\
& & P4 & 0.7719 & 0.7795 & 0.7879 & \colordiff{-0.0076} & \colordiff{-0.0160}$^\dag$ \\
& & P5 & 0.8291 & 0.8518 & 0.8526 & \colordiff{-0.0227}$^\dag$ & \colordiff{-0.0235}$^\dag$ \\
\midrule
\multirow{10}{*}{Bazinga!}
& \multirow{5}{*}{LLaMA}
& P1 & 0.6656 & 0.6718 & 0.6738 & \colordiff{-0.0062} & \colordiff{-0.0082} \\
& & P2 & 0.7284 & 0.7467 & 0.7371 & \colordiff{-0.0183}$^\dag$ & \colordiff{-0.0087} \\
& & P3 & 0.5274 & 0.5814 & 0.5750 & \colordiff{-0.0540}$^\dag$ & \colordiff{-0.0476}$^\dag$ \\
& & P4 & 0.6279 & 0.6432 & 0.6531 & \colordiff{-0.0153}$^\dag$ & \colordiff{-0.0252}$^\dag$ \\
& & P5 & 0.7537 & 0.8124 & 0.8175 & \colordiff{-0.0587}$^\dag$ & \colordiff{-0.0638}$^\dag$ \\
\cmidrule(lr){2-8}
& \multirow{5}{*}{Qwen}
& P1 & 0.7540 & 0.7549 & 0.7663 & \colordiff{-0.0009} & \colordiff{-0.0123}$^\dag$ \\
& & P2 & 0.7396 & 0.7588 & 0.7626 & \colordiff{-0.0192}$^\dag$ & \colordiff{-0.0230}$^\dag$ \\
& & P3 & 0.7249 & 0.7452 & 0.7484 & \colordiff{-0.0203}$^\dag$ & \colordiff{-0.0235}$^\dag$ \\
& & P4 & 0.7129 & 0.7270 & 0.7415 & \colordiff{-0.0141}$^\dag$ & \colordiff{-0.0286}$^\dag$ \\
& & P5 & 0.7747 & 0.7774 & 0.7859 & \colordiff{-0.0027} & \colordiff{-0.0112}$^\dag$ \\
\bottomrule
\end{tabular}
\end{table*}

\section{Two-Stage LLM-as-a-Judge Protocol for Response Generation}

To complement BERTScore with a more pragmatic evaluation, we adopt a two-stage LLM-as-a-judge protocol for the response generation task. In the first stage, the judge evaluates whether the generated response is appropriate given the dialogue history up to the target turn. In the second stage, the judge additionally observes the real future dialogue and evaluates whether the generated response remains compatible with the subsequent development of the conversation.

We use three judge models: Qwen3-32B, gpt-oss-120b, and Llama-3.3-70B-Instruct, and report the average score across them. In the pre-context stage, the judge rates four dimensions: \textit{Context Relevance}, \textit{Role Consistency}, \textit{Naturalness}, and \textit{Appropriateness}. In the post-context stage, the judge rates \textit{Future Coherence} and \textit{Non-Contradiction}. Each dimension is scored on a 1--5 scale, and we combine the two stage-level averages into a final weighted score (FWS):
\[
\text{FWS} = 0.6 \times \text{PreAvg} + 0.4 \times \text{PostAvg}.
\]

\noindent\textbf{Pre-context judge prompt.}

\begin{quote}\small
You are a very strict dialogue evaluation expert. Please evaluate the response generated by "\texttt{<speaker>}" based on the dialogue history up to now and the character relationships, using extremely rigorous standards.

Please rate from the following four dimensions (1 to 5 points for each dimension):
- 5 points: Flawless. Highly insightful, perfectly consistent with character traits and context, like the best possible reaction from a real human.\\
- 4 points: Excellent. Generally consistent with context and character, but with minor flaws.\\
- 3 points: Passing. Semantically smooth but somewhat mediocre, lacking distinctiveness, or just a perfunctory reply. If the tone does not reflect the character's traits, the maximum score is 3.\\
- 2 points: Poor. Obvious flaws, slightly off-topic, or the character's tone feels wrong.\\
- 1 point: Very Poor. Irrelevant, contradictory, or severely breaks character settings and relationships.

1. Context Relevance: Does the response closely and coherently follow the current dialogue context?\\
2. Role Consistency: Does the response highly align with the character's identity and their relationship with the listener?\\
3. Naturalness: Is the response natural and fluent in oral speech?\\
4. Appropriateness: Is the response reasonable and appropriate in the current situation?

\textbf{\#\#\# Characters and Relationships \#\#\#}\\
\texttt{<all\_chars>}
\texttt{<relation\_list>}

\textbf{\#\#\# Dialogue Up to Now \#\#\#}\\
\texttt{<dialogue>}

Just now "\texttt{<last\_speaker>}" said: "\texttt{<utterance>}"

Generated Response (said by "\texttt{<speaker>}"):\\
"\texttt{<pred\_reply>}"

Please strictly output the rating results in JSON format:
\begin{verbatim}
{
  "ContextRelevance": <1-5>,
  "RoleConsistency": <1-5>,
  "Naturalness": <1-5>,
  "Appropriateness": <1-5>
}
\end{verbatim}
\end{quote}

\noindent\textbf{Post-context judge prompt.}

\begin{quote}\small
You are a very strict dialogue logic evaluation expert. Please evaluate whether the response generated by "\texttt{<speaker>}" would lead to inconsistencies or contradictions in the complete context, including the real dialogue that occurs after the generated response.

Please rate from the following two dimensions (1 to 5 points for each dimension):
- 5 points: Flawless. The generated response not only does not contradict subsequent dialogue but even perfectly facilitates the development of the following plot.\\
- 4 points: Excellent. No obvious contradictions, smoothly links to subsequent dialogue.\\
- 3 points: Passing. While there are no serious contradictions, the transition is slightly stiff, or it is a throwaway line.\\
- 2 points: Poor. Partial logical conflict with subsequent events, making the following dialogue feel abrupt.\\
- 1 point: Very Poor. Severe contradiction; after this sentence is generated, the future dialogue becomes logically impossible.

1. Future Coherence: Can the generated response smoothly lead to the subsequent real dialogue?\\
2. Non-Contradiction: Does the generated response conflict with facts or emotions revealed in the subsequent dialogue?

\textbf{\#\#\# Characters and Relationships \#\#\#}\\
\texttt{<all\_chars>}
\texttt{<relation\_list>}

\textbf{\#\#\# Dialogue Up to Now \#\#\#}\\
\texttt{<dialogue>}

Just now "\texttt{<last\_speaker>}" said: "\texttt{<utterance>}"

Generated Response (said by "\texttt{<speaker>}", note that this is not part of the real dialogue):\\
"\texttt{<pred\_reply>}"

\textbf{\#\#\# Real Dialogue Occurring After This \#\#\#}\\
\texttt{<future\_dialogue>}

Please strictly output the rating results in JSON format:
\begin{verbatim}
{
  "FutureCoherence": <1-5>,
  "NonContradiction": <1-5>
}
\end{verbatim}
\end{quote}

\section{Prompt Template for Addressee Mention Type Classification}
\label{app: mention type}

To support the mention-type analysis in the main paper, we annotate each utterance with an addressee mention type using three LLM annotators: gpt-oss-120b, LLaMA-3.1-8B, and Qwen3-32B. For each instance, the final label is determined by majority vote across the three models. In cases where the three annotators do not reach a majority consensus or where the prediction is judged to be ambiguous, the final label is manually adjudicated by a human annotator.

The classification task is formulated as a five-way classification over the following categories: \textbf{Name-based Reference}, \textbf{Role-based Reference}, \textbf{Implicit Reference}, \textbf{Non-addressee Reference}, and \textbf{No-reference}. Below we provide the translated prompt used for this annotation task.

\noindent\textbf{Category definitions with examples.}

\begin{quote}\small
\textbf{A. Name-based Reference}\\
The speaker directly addresses the addressee by name or nickname.\\
Example: ``Zhengpeng, did you finish your homework?''

\vspace{0.5em}
\textbf{B. Role-based Reference}\\
The speaker refers to the addressee using a role title or social role.\\
Example: ``Mom, can you help me with this?''

\vspace{0.5em}
\textbf{C. Implicit Reference}\\
The speaker does not mention the addressee's name or role, but the utterance still implicitly refers to the addressee, for example through second-person expressions or discourse cues.\\
Example: ``Can you stop doing that?''

\vspace{0.5em}
\textbf{D. Non-addressee Reference}\\
The speaker mentions a person, but that reference does not refer to the intended addressee.\\
Example: ``I'm curious about what Xiaobao did today?''

\vspace{0.5em}
\textbf{E. No-reference}\\
The utterance does not contain any explicit or implicit reference to any person.\\
Example: ``It's getting late.''
\end{quote}

\noindent\textbf{Prompt template.}

\begin{quote}\small
You are an efficient and precise dialogue analysis AI. Your task is to classify an utterance based on how the speaker refers to people in the dialogue, with a focus on addressee identification. There are five possible categories: A, B, C, D, and E.

\textbf{A}: Directly addresses the addressee by name or nickname.\\
\textbf{B}: Refers to the addressee using a role title.\\
\textbf{C}: Does not mention the addressee's name or role, but still implicitly refers to the addressee.\\
\textbf{D}: Uses a linguistic reference, but it refers to someone other than the addressee.\\
\textbf{E}: Does not use any explicit or implicit linguistic reference to any person.

Please respond with a single uppercase letter. For example, if the correct answer is A, reply: \texttt{Answer: A}

You will be provided with the character list, their relations, the dialogue context, and a specific utterance. You are also given the intended addressee.

\textbf{\#\#\# Character Info \#\#\#}\\
Character List: \texttt{<all characters>}\\
\texttt{<relation\_list>}

\textbf{\#\#\# Dialogue \#\#\#}\\
\texttt{<full dialogue context>}

\textbf{\#\#\# Task \#\#\#}\\
Determine whether the utterance "\texttt{<utterance>}" spoken by \texttt{<speaker>} to \texttt{<addressee>} belongs to Type A, B, C, D, or E.\\
Only reply with one of the following letters: \texttt{A}, \texttt{B}, \texttt{C}, \texttt{D}, or \texttt{E}.
\end{quote}

\section{Additional Analyses}

\subsection{Effect of Name Anonymization}

To assess whether models rely on lexical identity cues, we anonymize all character names in MPDD and Bazinga! using neutral identifiers (\textit{P1, P2, P3, ...}) while preserving the dialogue context and social-information blocks. This analysis allows us to examine whether the benefit of participant-level social information persists when explicit name cues are removed.

Table~\ref{tab:alias-delta-full} shows that the effect of anonymization is model- and dataset-dependent on addressee recognition task. In particular, removing lexical identity cues does not uniformly reduce the usefulness of social information. Instead, the results suggest that different models rely on names and structured social metadata to different degrees.
\begin{table*}[t]
\centering
\small
\setlength{\tabcolsep}{4pt}
\caption{Exact Match Accuracy under the \textit{Name} and \textit{Alias} settings.}
\label{tab:alias-delta-full}
\begin{tabular}{lcccc|cccc}
\toprule
\textbf{Model} &
\multicolumn{4}{c|}{\textbf{Name}} &
\multicolumn{4}{c}{\textbf{Alias}} \\
\cmidrule(lr){2-5} \cmidrule(lr){6-9}
& \textbf{w/o} & \textbf{Pos.} & \textbf{Rel.} & $\Delta_{\text{Pos.}}$ / $\Delta_{\text{Rel.}}$
& \textbf{w/o} & \textbf{Pos.} & \textbf{Rel.} & $\Delta_{\text{Pos.}}$ / $\Delta_{\text{Rel.}}$ \\
\midrule
\multicolumn{9}{l}{\textbf{MPDD}} \\
Qwen  & 0.731 & 0.739 & 0.739 & \colordiff{-0.008}/\colordiff{-0.008} & 0.594 & 0.606 & 0.621 & \colordiff{-0.012} / \colordiff{-0.027} \\
LLaMA & 0.614 & 0.629 & 0.641 & \colordiff{-0.015}/\colordiff{-0.027}  & 0.561 & 0.579 & 0.581 & \colordiff{-0.018} / \colordiff{-0.020} \\
\midrule
\multicolumn{9}{l}{\textbf{Bazinga!}} \\
Qwen  & 0.628 & 0.669 & 0.664 & \colordiff{-0.041}/\colordiff{-0.036}  & 0.528 & 0.538 & 0.579 & \colordiff{-0.010} / \colordiff{-0.051} \\
LLaMA & 0.642 & 0.726 & 0.724 & \colordiff{-0.084}/\colordiff{-0.082}  & 0.622 & 0.639 & 0.651 & \colordiff{-0.017} / \colordiff{-0.029} \\
\bottomrule
\end{tabular}
\end{table*}

\subsection{Next Speaker Prediction}

As an additional analysis, we explore next speaker prediction (NSP). Unlike the turn-taking judgement task, which asks separately for each candidate participant whether speaking next is plausible, NSP requires the model to select the next speaker from all participants. It is therefore a substantially more open-ended and uncertain task.

We include NSP only as a supplementary discussion, since multiple participants may plausibly continue the dialogue at the same point. Nevertheless, it remains relevant because addressee recognition and next-speaker behavior are closely related. In MPDD, the next speaker is included in the current utterance's addressee set in 77.7\% of cases, and in Bazinga! the corresponding proportion is 81.3\%.

As shown in Table~\ref{tab:nsp_results}, NSP is considerably harder and the gains from social information are smaller and less stable than in the main tasks. Still, \textsc{Relation} information often provides modest improvements over the no-information baseline, suggesting that social cues can also support NSP despite the high uncertainty.

\begin{table*}[h]
\centering
\small
\setlength{\tabcolsep}{5pt}
\caption{Supplementary results for next speaker prediction (NSP). $\Delta_{\text{Pos}}$ and $\Delta_{\text{Rel}}$ denote differences from the no-information baseline. Results show that social information still provides modest gains in many settings.}
\label{tab:nsp_results}
\begin{tabular}{lccccccccc}
\toprule
\multirow{2}{*}{\textbf{Model}} &
\multicolumn{4}{c}{\textbf{MPDD}} &&
\multicolumn{4}{c}{\textbf{Bazinga!}} \\
\cmidrule(lr){2-5} \cmidrule(lr){7-10}
& \textbf{w/o} & \textbf{Pos.} & \textbf{Rel.} & $\Delta_{\text{Pos.}}$ / $\Delta_{\text{Rel.}}$
&& \textbf{w/o} & \textbf{Pos.} & \textbf{Rel.} & $\Delta_{\text{Pos.}}$ / $\Delta_{\text{Rel.}}$ \\
\midrule
GPT
& 0.547 & 0.578 & 0.573 & \colordiff{-0.031} / \colordiff{-0.027}
&& 0.571 & 0.622 & 0.617 & \colordiff{-0.051} / \colordiff{-0.046} \\
Qwen
& 0.537 & 0.522 & 0.535 & \colordiff{0.015} / \colordiff{0.002}
&& 0.547 & 0.555 & 0.557 & \colordiff{-0.008} / \colordiff{-0.010} \\
LLaMA
& 0.378 & 0.380 & 0.379 & \colordiff{-0.002} / \colordiff{-0.001}
&& 0.371 & 0.393 & 0.384 & \colordiff{-0.022} / \colordiff{-0.013} \\
GPT-OSS
& 0.565 & 0.588 & 0.596 & \colordiff{-0.023} / \colordiff{-0.031}
&& 0.600 & 0.645 & 0.643 & \colordiff{-0.045} / \colordiff{-0.043} \\
\bottomrule
\end{tabular}
\end{table*}

\subsection{Comparison with Fine-tuned Encoder Baselines}

We additionally compare against fine-tuned encoder baselines on addressee recognition task, using RoBERTa on MPDD and BERT on Bazinga!, under the same \textsc{None}, \textsc{Position}, and \textsc{Relation} settings. Both models benefit substantially from participant-level social information. On MPDD, RoBERTa improves from 0.5999 to 0.6123 / 0.6422 in Exact Match Accuracy and from 0.7736 to 0.8070 / 0.8226 in F1. On Bazinga!, BERT improves from 0.5184 to 0.5860 / 0.6486 in Exact Match Accuracy and from 0.6800 to 0.7279 / 0.7864 in F1. In both cases, \textsc{Relation} provides the strongest gains. This supports our broader conclusion that social information is useful beyond prompt-based LLM settings.

\begin{table*}[h]
\centering
\small
\setlength{\tabcolsep}{3.5pt}
\caption{Fine-tuned encoder baselines under different social-information settings. Diff $=$ (w/o $-$ w/ Info).}
\label{tab:encoder_baselines}
\begin{tabular}{llcccccc}
\toprule
Dataset & Model & Metric & w/o & Pos. & Rel. & Diff Pos. & Diff Rel. \\
\midrule
\multirow{2}{*}{MPDD}
& \multirow{2}{*}{RoBERTa}
& Exact Match & 0.5999 & 0.6123 & 0.6422 & \colordiff{-0.0124}$^\dag$ & \colordiff{-0.0423}$^\dag$ \\
& & F-score & 0.7736 & 0.8070 & 0.8226 & \colordiff{-0.0334}$^\dag$ & \colordiff{-0.0489}$^\dag$ \\
\midrule
\multirow{2}{*}{Bazinga!}
& \multirow{2}{*}{BERT}
& Exact Match & 0.5184 & 0.5860 & 0.6486 & \colordiff{-0.0676}$^\dag$ & \colordiff{-0.1303}$^\dag$ \\
& & F-score & 0.6800 & 0.7279 & 0.7864 & \colordiff{-0.0479}$^\dag$ & \colordiff{-0.1064}$^\dag$ \\
\bottomrule
\end{tabular}
\end{table*}

\subsection{Multimodal Extension on Bazinga!}

Our main study focuses on text-based addressee recognition. We view this setting as both practically relevant and methodologically important: many real-world scenarios, such as chat logs, meeting transcripts, and subtitle-style corpora, contain only textual dialogue, and text-only evaluation also provides a necessary baseline for future multimodal systems.

At the same time, we agree that real-world addressee recognition can depend on non-verbal cues, such as gaze, speaker orientation, or broader spatial context. To explore whether our main findings extend beyond the text-only setting, we conduct a preliminary text+audio pilot on Bazinga!, which provides aligned audio. Specifically, we parse utterance-level timestamps and extract the corresponding audio segments for each turn, yielding 1602 evaluation samples.

In this pilot setting, structured social information remains beneficial even when audio is included. For example, with Qwen2.5-Omni, Exact Match Accuracy improves from 0.6754 to 0.7116 under the \textsc{Position} setting, while F1 improves from 0.7628 to 0.8028. The \textsc{Relation} setting also improves over the no-information baseline, reaching 0.7022 in Exact Match Accuracy and 0.7966 in F1. These preliminary results suggest that the usefulness of participant-level social cues is not merely an artifact of text-only prompting; rather, social information appears to provide complementary value even when audio is available.

We emphasize, however, that this multimodal extension is only a pilot study. Richer multimodal signals, such as gaze and spatial context, remain beyond the scope of the present work and are left for future research.

\begin{table*}[h]
\centering
\small
\setlength{\tabcolsep}{3.5pt}
\caption{Preliminary text+audio results on Bazinga! (1602 samples). Diff $=$ (w/o $-$ w/ Info).}
\label{tab:multimodal_bazinga}
\begin{tabular}{llcccccc}
\toprule
Model & Metric & w/o & Pos. & Rel. & Diff Pos. & Diff Rel. \\
\midrule
\multirow{2}{*}{Qwen2.5-Omni}
& Exact Match & 0.6754 & 0.7116 & 0.7022 & \colordiff{-0.0362} & \colordiff{-0.0268} \\
& F-score & 0.7628 & 0.8028 & 0.7966 & \colordiff{-0.0400} & \colordiff{-0.0338} \\
\bottomrule
\end{tabular}
\end{table*}

\newtcolorbox{qualbox}[1]{
  enhanced,
  breakable,
  colback=gray!18,
  colframe=gray!55,
  boxrule=0.55pt,
  arc=1.5mm,
  left=1.2mm,
  right=1.2mm,
  top=1.0mm,
  bottom=1.0mm,
  title={#1},
  fonttitle=\bfseries,
  coltitle=black
}

\subsection{Fine-tuned LLMs for Addressee Recognition}
\label{app:finetune}

\paragraph{Setup.}
We fine-tune LLaMA-3.1-8B-Instruct with QLoRA (4-bit NF4 base weights) and
Qwen3-32B (non-thinking mode) with LoRA on bf16 weights for addressee recognition.
Inputs are identical to the zero-shot prompts (Appendix~\ref{app:prompts}), 
with the social-information block set to NONE, POSITION, or RELATION;
the training loss is computed on the answer tokens only.
Following the encoder baselines (Appendix~H.3), we use 5-fold cross-validation
with dialogue-level splits, so that no dialogue appears in both the training and
test folds, and pool predictions over all folds.
Every utterance is therefore scored exactly once, on the same set as the zero-shot
results in Table~\ref{tab:social-info-impact}.
We use LoRA rank $r=16$, $\alpha=32$, dropout 0.05 on all attention and MLP
projections, learning rate $2\times10^{-4}$ (LLaMA) / $1\times10^{-4}$ (Qwen) with
cosine decay and 3\% warm-up, 2 epochs, effective batch size 16, maximum length
3{,}072 tokens (no example exceeded it), seed 42, and greedy decoding at test time.
Each run used one NVIDIA H100 (80GB); the 60 runs
(2 models $\times$ 2 datasets $\times$ 3 settings $\times$ 5 folds)
took about 42 GPU hours in total.
Significance is computed with a paired $t$-test on per-utterance scores.

\paragraph{Results.}
Table~\ref{tab:finetune} shows three findings.
First, fine-tuning substantially improves over zero-shot prompting in every setting
(9.4--20.2 points in Exact Match without social information).
Second, social information remains beneficial after fine-tuning:
both POSITION and RELATION significantly outperform NONE for both models on both datasets.
Third, the advantage of RELATION over POSITION seen in the zero-shot setting
largely disappears after fine-tuning: the two differ significantly only for LLaMA,
where RELATION is slightly better on MPDD in F-score and POSITION is better on
Bazinga! in both metrics.
Once the model is trained on the task, coarse role information appears to carry
much of the useful signal; on Bazinga!, whose relations were inferred from
external resources (Appendix~\ref{tab:bazinga_construct}), 
the coarser POSITION labels may also be more robust to annotation noise.

\begin{table*}[t]
    \centering
    \setlength{\tabcolsep}{2.7pt}
    \small
    \begin{tabular}{clcccccccccc}
    \toprule
    \multirow{3}{*}{\textbf{Dataset}}
    & \multirow{3}{*}{\textbf{Model}}
    & \multicolumn{4}{c}{\textbf{Exact Match Accuracy}}
    & & \multicolumn{4}{c}{\textbf{F-score}} \\
    \cmidrule(lr){3-6}
    \cmidrule(lr){8-11}
    &
    & \textbf{w/o}
    & \multicolumn{2}{c}{\textbf{w/ Info}}
    & \textbf{Diff}
    &
    & \textbf{w/o}
    & \multicolumn{2}{c}{\textbf{w/ Info}}
    & \textbf{Diff} \\
    \cmidrule(lr){4-5}
    \cmidrule(lr){9-10}
    &
    & \textbf{Info}
    & \textbf{Pos.}
    & \textbf{Rel.}
    & \textbf{(w/o - w/)}
    &
    & \textbf{Info}
    & \textbf{Pos.}
    & \textbf{Rel.}
    & \textbf{(w/o - w/)} \\
    \midrule

    \multirow{2}{*}{\textbf{MPDD}}
    & Qwen
    & 0.8249 & 0.8853 & 0.8876
    & \colordiff{-0.0604}$^\dag$ / \colordiff{-0.0627}$^\dag$
    &&
    0.9121 & 0.9502 & 0.9512
    & \colordiff{-0.0381}$^\dag$ / \colordiff{-0.0391}$^\dag$ \\

    &
    LLaMA
    & 0.8089 & 0.8679 & 0.8728
    & \colordiff{-0.0590}$^\dag$ / \colordiff{-0.0639}$^\dag$
    &&
    0.8980 & 0.9386 & 0.9424$^\ddag$
    & \colordiff{-0.0406}$^\dag$ / \colordiff{-0.0444}$^\dag$ \\

    \midrule

    \multirow{2}{*}{\textbf{Bazinga!}}
    & Qwen
    & 0.8299 & 0.9034 & 0.9007
    & \colordiff{-0.0734}$^\dag$ / \colordiff{-0.0707}$^\dag$
    &&
    0.8733 & 0.9404 & 0.9395
    & \colordiff{-0.0671}$^\dag$ / \colordiff{-0.0661}$^\dag$ \\

    &
    LLaMA
    & 0.8152 & 0.9061$^\ddag$ & 0.8960
    & \colordiff{-0.0908}$^\dag$ / \colordiff{-0.0808}$^\dag$
    &&
    0.8615 & 0.9411$^\ddag$ & 0.9344
    & \colordiff{-0.0796}$^\dag$ / \colordiff{-0.0729}$^\dag$ \\

    \bottomrule
    \end{tabular}

    \caption{
    Addressee recognition with fine-tuned LLMs (5-fold dialogue-level cross-validation,
    pooled over all utterances; zero-shot results on the same utterances are in
    Table~\ref{tab:social-info-impact}). Definitions of Diff, color coding, and significance
    notation follow Table~\ref{tab:social-info-impact}. $^\ddag$: significantly better than the
    other social-information setting ($p<0.05$). Every fine-tuned score is significantly higher
    than the corresponding zero-shot score in Table~\ref{tab:social-info-impact} ($p<0.05$).
    }
    \label{tab:finetune}
\end{table*}

\section{Response Generation}
\label{app:response-generation}

\subsection{Agreement Between LLM Judges}
\label{app:judge-agreement}

Table~\ref{tab:judge-agreement} reports inter-judge agreement
across 95,340 MPDD and 28,092 Bazinga! replies.
Pairwise correlations are moderate ($r = 0.51$--$0.68$),
while the three-judge average achieves higher reliability
(ICC(3,$k$) = 0.82 and 0.78, respectively).
The \textit{All three} rows report pairwise averages for
$r$ and $\rho$.
These results support our use of aggregated judge scores.

\begin{table}[h]
    \centering
    \small
    \setlength{\tabcolsep}{3pt}
    \begin{tabular}{lcccc}
    \toprule
    \textbf{Judges} & $\boldsymbol{r}$ & $\boldsymbol{\rho}$
    & \textbf{ICC(3,1)} & \textbf{ICC(3,$k$)} \\
    \midrule
    \multicolumn{5}{l}{\textbf{MPDD}} \\
    Qwen -- GPT-OSS  & 0.580 & 0.526 & 0.559 & -- \\
    Qwen -- LLaMA    & 0.679 & 0.629 & 0.679 & -- \\
    GPT-OSS -- LLaMA & 0.575 & 0.517 & 0.551 & -- \\
    \textit{All three} & 0.611 & 0.558 & 0.603 & \textbf{0.820} \\
    \midrule
    \multicolumn{5}{l}{\textbf{Bazinga!}} \\
    Qwen -- GPT-OSS  & 0.512 & 0.443 & 0.487 & -- \\
    Qwen -- LLaMA    & 0.622 & 0.538 & 0.611 & -- \\
    GPT-OSS -- LLaMA & 0.534 & 0.446 & 0.530 & -- \\
    \textit{All three} & 0.556 & 0.476 & 0.546 & \textbf{0.783} \\
    \bottomrule
    \end{tabular}
    \caption{Agreement between LLM judges on FWS.
    $r$/$\rho$: Pearson/Spearman correlation.
    ICC(3,1)/ICC(3,$k$): single/average-measure consistency.}
    \label{tab:judge-agreement}
\end{table}

\subsection{Human Evaluation}
\label{app:human-eval}

To assess alignment between LLM and human judgments,
three annotators evaluated generated responses using the same
six criteria and 60/40 stage weighting as the LLM judges.
Each annotation unit contains one dialogue context and 12
replies from four models under three settings, presented
side by side.
We collected 2,400 ratings across 200 dialogue contexts
(1,200 ratings per dataset).
Attention-check items and ratings from annotators who failed
the checks or exhibited outlying rating patterns were excluded
from the analysis.

\paragraph{Human--judge agreement.}
Human scores correlate moderately with the three-judge average
(Spearman's $\rho = 0.56$, 95\% CI: 0.48--0.63;
Pearson's $r = 0.58$).
Confidence intervals are estimated by bootstrapping over
annotation units.

\paragraph{Human scores by setting.}
Table~\ref{tab:human-eval} shows that social information
improves human-rated scores for GPT and Qwen on MPDD, although only the GPT--\textsc{Relation}
gain is statistically significant.
The benefits of \textsc{Relation} are consistent with the
judge-based results, while the effects of \textsc{Position}
differ.
No significant differences are observed on Bazinga!.
Significance is assessed using paired $t$-tests over
(annotator, annotation unit) pairs.

\begin{table}[h]
    \centering
    \setlength{\tabcolsep}{3pt}
    \small
    \begin{tabular}{clcccc}
    \toprule
    \multirow{3}{*}{\textbf{Dataset}} &
    \multirow{3}{*}{\textbf{Model}} &
    \multicolumn{4}{c}{\textbf{Human Score}} \\
    \cmidrule(lr){3-6}
    & & \textbf{w/o}
    & \multicolumn{2}{c}{\textbf{w/ Info}}
    & \textbf{Diff} \\
    \cmidrule(lr){4-5}
    & & \textbf{Info} & \textbf{Pos.} & \textbf{Rel.}
    & \textbf{(w/o - w/)} \\
    \midrule
    \multirow{4}{*}{\textbf{MPDD}}
     & GPT & 3.07 & 3.46 & 3.48
     & \colordiff{-0.391}/\colordiff{-0.408}$^\dag$ \\
     & Qwen & 3.43 & 3.74 & 3.76
     & \colordiff{-0.315}/\colordiff{-0.328} \\
     & LLaMA & 2.67 & 2.67 & 2.70
     & \colordiff{-0.005}/\colordiff{-0.037} \\
     & GPT-OSS & 3.34 & 3.41 & 3.53
     & \colordiff{-0.070}/\colordiff{-0.195} \\
    \midrule
    \multirow{4}{*}{\textbf{Bazinga!}}
     & GPT & 3.27 & 3.44 & 3.21
     & \colordiff{-0.174}/\colordiff{0.055} \\
     & Qwen & 3.50 & 3.36 & 3.54
     & \colordiff{0.138}/\colordiff{-0.044} \\
     & LLaMA & 3.10 & 3.22 & 3.31
     & \colordiff{-0.125}/\colordiff{-0.216} \\
     & GPT-OSS & 3.42 & 3.56 & 3.44
     & \colordiff{-0.134}/\colordiff{-0.021} \\
    \bottomrule
    \end{tabular}
    \caption{Human evaluation of response generation
    (3 annotators, 2,400 ratings).
    Scores follow the same 1--5 scale and weighting as FWS.
    Diff, color coding, and significance notation follow
    Table~\ref{tab:response-generation}.}
    \label{tab:human-eval}
\end{table}

\subsection{Role Consistency}
\label{app:judge-dimensions}

Table~\ref{tab:role-consistency} reports the role-consistency
criterion, which measures whether a reply aligns with the
speaker's identity and relationship with the listener.
\textsc{Relation} improves role consistency across all models,
with larger gains on MPDD (0.14--0.23 points).
These gains exceed the corresponding FWS gains on MPDD,
suggesting a particular benefit for role consistency.
In contrast, \textsc{Position} yields mixed effects and
consistently underperforms \textsc{Relation}.

\begin{table}[h]
    \centering
    \setlength{\tabcolsep}{3pt}
    \small
    \begin{tabular}{clcccc}
    \toprule
    \multirow{3}{*}{\textbf{Dataset}} &
    \multirow{3}{*}{\textbf{Model}} &
    \multicolumn{4}{c}{\textbf{Role Consistency}} \\
    \cmidrule(lr){3-6}
    & & \textbf{w/o}
    & \multicolumn{2}{c}{\textbf{w/ Info}}
    & \textbf{Diff} \\
    \cmidrule(lr){4-5}
    & & \textbf{Info} & \textbf{Pos.} & \textbf{Rel.}
    & \textbf{(w/o - w/)} \\
    \midrule
    \multirow{4}{*}{\textbf{MPDD}}
     & GPT & 4.07 & 4.01 & 4.29
     & \colordiff{0.056}/\colordiff{-0.226} \\
     & Qwen & 4.21 & 4.18 & 4.41
     & \colordiff{0.023}/\colordiff{-0.202} \\
     & LLaMA & 3.35 & 3.46 & 3.53
     & \colordiff{-0.110}/\colordiff{-0.179} \\
     & GPT-OSS & 4.12 & 4.02 & 4.27
     & \colordiff{0.109}/\colordiff{-0.144} \\
    \midrule
    \multirow{4}{*}{\textbf{Bazinga!}}
     & GPT & 4.43 & 4.41 & 4.46
     & \colordiff{0.019}/\colordiff{-0.026} \\
     & Qwen & 4.31 & 4.32 & 4.39
     & \colordiff{-0.010}/\colordiff{-0.076} \\
     & LLaMA & 3.94 & 3.99 & 4.05
     & \colordiff{-0.050}/\colordiff{-0.108} \\
     & GPT-OSS & 4.47 & 4.50 & 4.56
     & \colordiff{-0.033}/\colordiff{-0.090} \\
    \bottomrule
    \end{tabular}
    \caption{Role-consistency scores (1--5) averaged over
    the three LLM judges.
    Diff and color coding follow
    Table~\ref{tab:response-generation}.}
    \label{tab:role-consistency}
\end{table}

\section{Qualitative Examples of Social-Cue Benefits}
\label{app:qualitative_cases}

To complement the quantitative results, we present two addressee recognition examples and one response generation example. MPDD examples are shown in English translation.

\small
\begin{qualbox}{Addressee Recognition: Position helps identify a role-based target}
\textbf{Translated utterance.} \textit{Zuo Zhengpeng: ``Students, I am sorry for being late today. Please turn to page 45. Let us begin a new lesson.''}

\textbf{Gold addressee.} Xu Lihua, Principal Sun

\textbf{Position information.} Xu Lihua is Zuo Zhengpeng's Inferior; Principal Sun is Zuo Zhengpeng's Superior

\textbf{Predictions.}
\begin{quote}\footnotesize
\textsc{None}: no prediction \\
\textsc{Position}: Xu Lihua
\end{quote}

\textbf{Interpretation.}
The utterance addresses \textit{students} rather than naming a specific person. With role information, the model can at least recover a student listener from the candidate set, whereas the no-information setting fails to map the role expression to any listed participant.
\end{qualbox}

\vspace{0.45em}

\begin{qualbox}{Addressee Recognition: Relation resolves an implicit family reference}
\textbf{Translated utterance.} \textit{Mother Chen: ``A-Xin, have a bite. You cannot skip dinner like this.''}

\textbf{Note.} In this storyline, ``A-Xin'' is Mother Chen's nickname for Zhang Ai; this is a different character from Zhang Xin (also called ``A-Xin'') in the next example.

\textbf{Gold addressee.} Zhang Ai

\textbf{Relation information.} Mother Chen is Zhang Ai's mother; Zhang Ai is her daughter

\textbf{Predictions.}
\begin{quote}\footnotesize
\textsc{None}: Mother Chen \\
\textsc{Relation}: Zhang Ai
\end{quote}

\textbf{Interpretation.}
The utterance uses an intimate family-style form of address rather than an explicit full name. Without relational cues, the model confuses the target with the speaker, while the relation-aware setting correctly recovers the mother-daughter link.
\end{qualbox}

\vspace{0.5em}

\begin{qualbox}{Response Generation: 3-way comparison on the same utterance}
\textbf{Translated context.}
\begin{quote}\footnotesize
Zhang Ai: ``A-Xin, come out. Officer Hu wants to talk to us about that matter.'' \\
Zhang Xin: ``Okay. I am coming right out.'' \\
Song Qiao: ``What is it?'' \\
Zhang Xin: ``Nothing important. It is about work. Mom, I am heading out for a bit.'' \\
Song Qiao: ``Come back early for dinner.'' \\
Zhang Xin: ``Okay. I am leaving now.''
\end{quote}

\textbf{Relation information.} Song Qiao is Zhang Xin's mother

\textbf{Translated target utterance.} \textit{Song Qiao: ``Be careful on the way.''}

\textbf{Generated continuation.}
\begin{quote}\footnotesize
\textsc{None}: ``Okay, I will be careful. Thank you, Brother Qiao. Hope everything goes well for you too.'' \\[0.3em]
\textsc{Position}: ``Okay, Song Qiao, I will be careful. Please take care as well.'' \\[0.3em]
\textsc{Relation}: ``Okay, Mom, I will be careful. I will be back in time for dinner.''
\end{quote}

\textbf{Interpretation.}
This example shows the clearest qualitative contrast among the three settings. Without social information, the continuation uses an implausible address term; \textsc{Position} improves formality but still sounds distant; \textsc{Relation} produces the most natural continuation by recovering the mother-daughter relationship and linking it back to the dinner context.
\end{qualbox}

\normalsize

\section{GenAI Usage Disclosure}

We utilized generative AI tools solely for grammar and language refinement. 
All content was reviewed and edited by the author(s), who take full responsibility for the final manuscript.

\end{document}